\documentclass[10pt,twocolumn,letterpaper]{article}

\usepackage{cvpr}              

\usepackage{longtable}
\usepackage{tikz}
\usetikzlibrary{positioning,calc}
\usepackage{tabularx}
\usepackage[para]{threeparttable}
\usepackage[x11names,dvipsnames,table]{xcolor}
\usepackage{makecell}
\usepackage{float}
\usepackage{graphicx}
\usepackage{subcaption}
\usepackage{multirow}
\newcolumntype{Y}{>{\raggedright\arraybackslash}X}
\usepackage[edges]{forest}

\usepackage[T1]{fontenc}
\usepackage[hyphens]{url}
\usepackage{xurl}

\usepackage{booktabs}
\usepackage[table]{xcolor}
\usepackage{amssymb} 
\usepackage{bm}
\usepackage{amsmath}
\usepackage{tcolorbox}
\tcbuselibrary{skins}
\usepackage{wrapfig}

\definecolor{cvprblue}{rgb}{0.21,0.49,0.74}
\usepackage[pagebackref,breaklinks,colorlinks,allcolors=cvprblue]{hyperref}

\def\paperID{8008} 
\def\confName{CVPR}
\def\confYear{2026}

\title{\vspace{-4pt}Med-CMR: A Fine-Grained Benchmark Integrating Visual Evidence and Clinical Logic for Medical Complex Multimodal Reasoning\vspace{-11pt}}

\author{
Haozhen Gong$^{1,\dagger}$ \quad
Xiaozhong Ji$^{2,\dagger}$ \quad
Yuansen Liu$^{1,\dagger}$ \quad
Wenbin Wu$^{1}$ \quad
Xiaoxiao Yan$^{4}$ \quad
Jingjing Liu$^{1}$
\\
Kai Wu$^{3}$ \quad
Jiazhen Pan$^{5}$ \quad
Bailiang Jian$^{5}$ \quad
Jiangning Zhang$^{6}$ \quad
Xiaobin Hu$^{1,*}$ \quad
Hongwei Bran Li$^{1}$
\\[3pt]
$^{1}$National University of Singapore \quad
$^{2}$Nanjing University \quad
$^{3}$Tongji University \quad
$^{4}$Ruijin Hospital
\\
$^{5}$Technical University of Munich \quad
$^{6}$Zhejiang University
\\[3pt]
$\dagger$ Equal contribution \quad * Corresponding author
}

\newcommand{\method}{Med-CMR}

\begin{document}
\maketitle
\begin{abstract}

MLLMs are beginning to appear in clinical workflows, but their ability to perform complex medical reasoning remains unclear. We present \method, a fine-grained \textbf{Med}ical \textbf{C}omplex \textbf{M}ultimodal \textbf{R}easoning benchmark. \method~distinguishes from existing counterparts by three core features: 1) \textbf{Systematic capability decomposition}, splitting medical multimodal reasoning into fine-grained visual understanding and multi-step reasoning to enable targeted evaluation; 2) \textbf{Challenging task design}, with visual understanding across three key dimensions (small-object detection, fine-detail discrimination, spatial understanding) and reasoning covering four clinically relevant scenarios (temporal prediction, causal reasoning, long-tail generalization, multi-source integration); 3) \textbf{Broad, high-quality data coverage}, comprising 20,653 Visual Question Answering (VQA) pairs spanning 11 organ systems and 12 imaging modalities, validated via a rigorous two-stage (human expert + model-assisted) review to ensure clinical authenticity. 
We evaluate 18 state-of-the-art MLLMs with \method, revealing GPT-5 as the top-performing commercial model: 57.81 accuracy on multiple-choice questions (MCQs) and a 48.70 open-ended score, outperforming Gemini 2.5 Pro (49.87 MCQ accuracy, 45.98 open-ended score) and leading open-source model Qwen3-VL-235B-A22B (49.34 MCQ accuracy, 42.62 open-ended score). However, specialized medical MLLMs do not reliably outperform strong general models, and long-tail generalization emerges as the dominant failure mode. Med-CMR thus provides a stress test for visual–reasoning integration and rare-case robustness in medical MLLMs, and a rigorous yardstick for future clinical systems. Project page: \href{https://github.com/LsmnBmnc/Med-CMR}{https://github.com/LsmnBmnc/Med-CMR}\vspace{-9pt}.


\end{abstract}

\section{Introduction}
\label{sec:intro}
Multimodal large language models (MLLMs) \cite{saab2024capabilities,moor2023med,li2023llava,chen2024huatuogpt,lin2025healthgpt,nath2024vila,tian2023chimed,chen2023huatuogpt,wang2025baichuan} are rapidly moving from proof-of-concept demos into tools that clinicians can actually touch. Before they are trusted in practice, we need to understand not only how often they are right, but how they reach decisions: can they detect subtle findings, integrate multiple images, track disease evolution, and reason about rare but critical scenarios? 
Concretely, we ask: \emph{to what extent can current MLLMs integrate medical images and clinical context to answer multi-step, reasoning-intensive questions, beyond basic VQA?}


Most existing multimodal medical benchmarks answer only a small part of this question ~\cite{bitton2023visit, zhu2024mmdocbench, li2025ddfav, ray2024ervqa, lim2024vlr, yu2023mm, yu2024mm, xu2023mmbench, lee2024vhelm, roberts2024image2struct,lau2018dataset,he2020pathvqa,zhang2023pmc,hu2024omnimedvqa, che2025lemon}. They are dominated by perception-level visual question answering, where the model describes an image or retrieves an obvious fact from a short context. This setup hides many of the hard cases that shape clinical decisions: tiny, low-contrast lesions; cross-modal comparisons; temporal change; causal chains linking symptoms, imaging, and outcomes; and long-tailed distribution in textbooks. As a result, today’s benchmarks provide limited visibility into the complex medical reasoning capabilities that matter in real workflows. 

Evaluating such capabilities requires three ingredients that current benchmarks largely lack. First, \textbf{systematic capability decomposition}: instead of treating ``medical multimodal reasoning" as a single score, we must separate fine-grained visual understanding from downstream reasoning, and further break both into clinically meaningful sub-dimensions. Second, \textbf{clinically aligned and deliberately challenging tasks}: questions should be built around real cases and explicitly target difficult settings such as temporal prediction, causal reasoning, long-tail generalization, and integration of multiple sources. Third, \textbf{broad and well-curated coverage} across organs, modalities, and disease processes, with expert review to ensure that questions remain realistic and clinically interpretable.

\begin{figure*}[htbp]
  \centering
  \includegraphics[width=\linewidth]{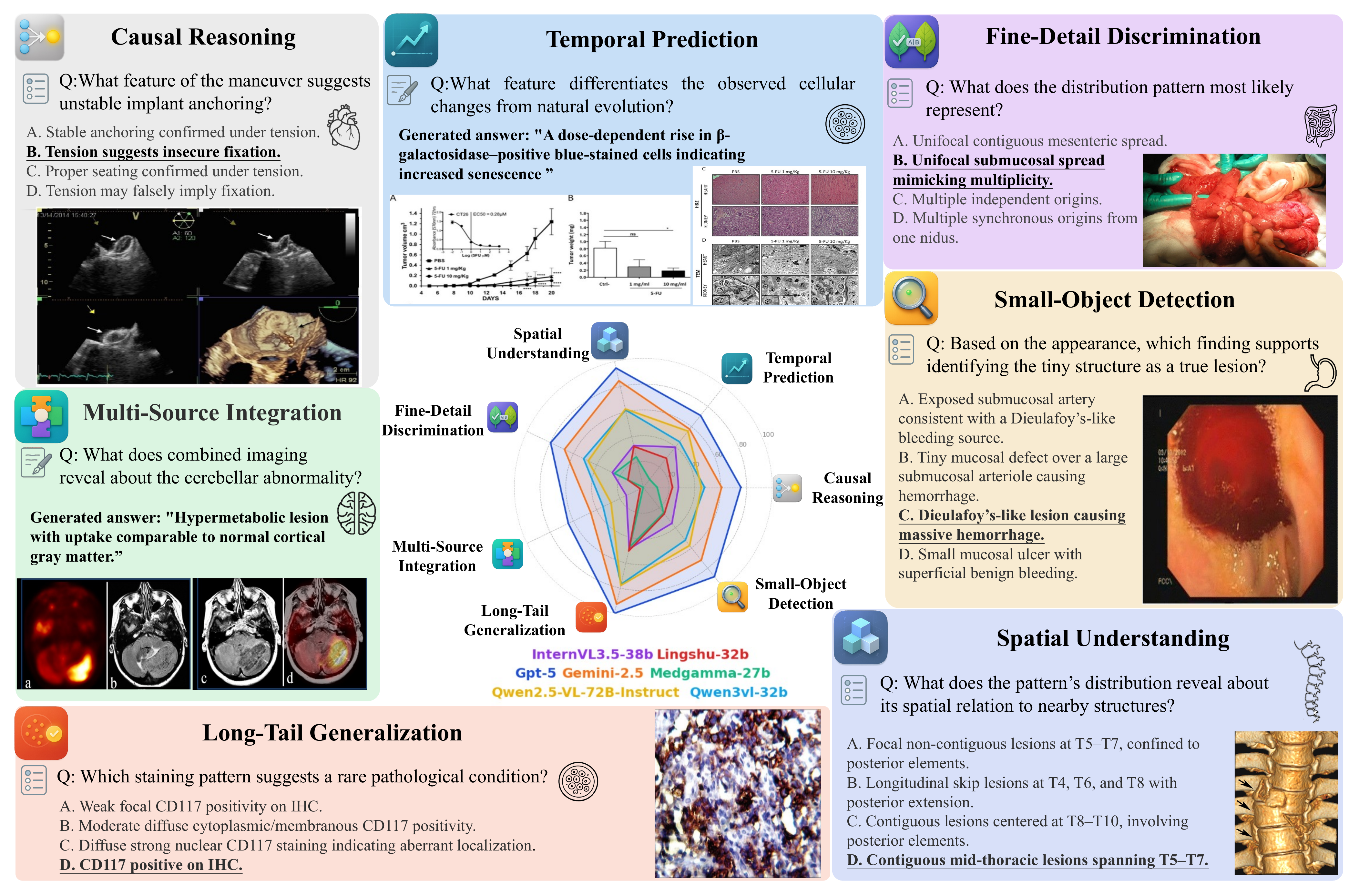}
  \vspace{-9mm}
  \caption{\textbf{Overview of Med-CMR.} Med-CMR decomposes medical multimodal reasoning complexity into visual complexity (i.e., small-object detection, fine-detail discrimination, and spatial understanding) and reasoning complexity (i.e., temporal prediction, causal reasoning, long-tail generalization, and multi-source integration). Each dimension corresponds to a specific task designed to evaluate the model’s capability in that dimension.}

  \label{fig:pipeline}
\end{figure*}

To address these gaps, we introduce \method, a comprehensive benchmark that systematically evaluates MLLMs across multiple dimensions. Specifically, we categorizes reasoning complexity into three visual dimensions, \textit{small-object detection, fine-detail discrimination, and spatial understanding}; and four reasoning complexity dimensions, \textit{temporal prediction, causal reasoning, long-tail generalization, and multi-source integration}. Each dimension corresponds to a distinct capability of MLLMs, enabling detailed diagnosis of model strengths and weaknesses. 
We collect data from authentic journal case reports and clinical research articles, followed by a two-stage filtering process combining human expert validation and model-assisted screening to ensure both quality and difficulty.
Ultimately, \method~achieves comprehensive coverage across 11 organ systems and 12 modalities, establishing a diverse foundation for assessing complex clinical reasoning~(Figure~\ref{fig:pipeline}).

During the evaluation phase, we assess models using both multiple-choice and open-ended questions.
While multiple-choice questions measure factual correctness, open-ended responses evaluate reasoning and generation quality.
The latter are graded by an external, standard-aligned LLM based on four complementary criteria: Consistency, Coherence, Visual accuracy, and Ground-truth correctness, offering a holistic view of model performance. As validated in Sec. \hyperref[sec:human]{6.1}, the LLM-based scoring shows high reliability and strong consistency with human judgment.

We evaluate 18 proprietary and open-source MLLMs on Med-CMR to systematically compare their specific capabilities in complex clinical reasoning. GPT-5 achieves the best overall performance, reaching 57.81\% accuracy on MCQs and a score of 48.70 on open-ended tasks. We conduct a detailed error analysis of GPT-5, the current upper-bound model, to identify the main causes of its failures and suggest directions for advancing real-world complex clinical reasoning. Through experiments, we analyze and validate the potential reasons why medical MLLMs do not consistently outperform general models in complex reasoning. 

Our main contributions are three-fold:
\begin{itemize}
    \item We propose Med-CMR, the first benchmark that enables fine-grained evaluation of clinical complex reasoning in MLLMs. Med-CMR decomposes reasoning into seven tasks that capture distinct sources of medical complexity, providing a structured and clinically grounded assessment of models’ capabilities.
    \item We design a unified evaluation protocol combining multiple-choice and open-ended questions, where the latter are graded by an external, standard-aligned LLM along consistency, coherence, visual accuracy, and ground-truth correctness. This framework yields a holistic view of both answer correctness and reasoning quality.
    \item We benchmark 18 proprietary and open-source vision–language models, including medical MLLMs and general-purpose MLLMs, and conduct detailed error and ablation analyses. We show that medical MLLMs do not consistently outperform general models in complex clinical reasoning, revealing concrete failure modes and directions for future improvement.

\end{itemize}
\begin{table*}[ht]
\centering
\scriptsize 
\setlength{\tabcolsep}{4pt} 
\renewcommand{\arraystretch}{1.1} 
\caption{Comparison of \method~with prior multimodal medical benchmarks. The background colors denote benchmark categories: purple for early Medical VQA datasets focusing on basic understanding, orange for benchmarks that start to touch complex reasoning, and light yellow for our \method~with fine-grained complex medical reasoning evaluation.}
\vspace{-3mm}
\label{tab:1}
\resizebox{\textwidth}{!}{
\begin{tabular}{
    l
    c
    c
    c
    c
    c
    c
    c
    c
}
\toprule
\textbf{Benchmark} & 
\textbf{\# Images} & 
\textbf{\# Size} & 
\textbf{QA Type} & 
\textbf{Annotation Type} &
\textbf{Broad Coverage} &
\textbf{Challenging Tasks} &
\textbf{Fine-grained Eval.} \\
\midrule

\rowcolor{blue!7}
VQA-RAD \cite{lau2018dataset} & 204 & 451 & Open-ended & Human & $\bm{\times}$ & $\bm{\times}$ & $\bm{\times}$ \\
\rowcolor{blue!7}
VQA-Med \cite{abacha2019vqa} & 500 & 500 & Open-ended & Automatic + Human & $\bm{\times}$ & $\bm{\times}$ & $\bm{\times}$ \\
\rowcolor{blue!7}
Path-VQA \cite{he2020pathvqa} & 858 & 6,719 & Open-ended & Automatic + Human & $\bm{\times}$ & $\bm{\times}$ & $\bm{\times}$ \\
\rowcolor{blue!7}
Slake-En \cite{liu2021slake} & 96 & 1,061 & Open-ended & Automatic + Human & $\bm{\times}$ & $\bm{\times}$ & $\bm{\times}$ \\

\rowcolor{blue!7}
PMC-VQA \cite{zhang2023pmc} & 29,021 & 33,430 & Open-ended & Automatic & $\bm{\checkmark}$ & $\bm{\times}$ & $\bm{\times}$ \\
\rowcolor{blue!7}
OmniMedVQA \cite{hu2024omnimedvqa} & 118,010 & 127,995 & MCQ & Automatic & $\bm{\checkmark}$ & $\bm{\times}$ & $\bm{\times}$ \\
\rowcolor{blue!7}
GMAI-MMBench \cite{ye2024gmai} & 21,180 & 21,281 & MCQ + MRQ & Automatic & $\bm{\checkmark}$ & $\bm{\times}$ & $\bm{\times}$ \\

\rowcolor{orange!10}
MedXpertQA MM \cite{zuo2025medxpertqa} & 2,852 & 2,000 & MCQ & Human & $\bm{\checkmark}$ & $\bm{\checkmark}$ & $\bm{\times}$ \\
\rowcolor{orange!10}
HIE-Reasoning \cite{baovisual} & 133 & 749 & MCQ + Open-ended & Human & $\bm{\times}$ & $\bm{\checkmark}$ & $\bm{\times}$ \\

\midrule
\rowcolor{yellow!10}
\textbf{\method} & 20,653 & 20,653 & MCQ + open-ended & Automatic + Human & $\bm{\checkmark}$ & $\bm{\checkmark}$ & $\bm{\checkmark}$ \\

\bottomrule
\end{tabular}
} 

\vspace{-4mm}
\end{table*}
\section{\method~}

We propose \method, a large-scale benchmark that evaluates  multimodal medical complex reasoning through fine-grained categorization of clinical complexity. In Sec.\hyperref[sec:definition]{2.1}, we introduce how we define and design complex medical questions to enable granular evaluation of MLLMs. Table \hyperref[tab:1]{1} compares \method~with prior multimodal medical benchmarks and presents more detailed information. Details of prior multimodal medical benchmarks are provided in the related work Section of Supplementary Material.

\subsection{How to Break Down Medical Reasoning Complexity and Ask Complex Questions?}
\label{sec:definition}
In clinical practice, perception and reasoning are closely connected: physicians interpret uncertain evidence, integrate information from multiple modalities, and make diagnostic decisions under both cognitive and perceptual constraints.
Medical reasoning is inherently demanding shaped by uncertainty, incomplete information, and the need to integrate multimodal evidence across time and context \cite{elstein2002clinical,eva2005clinical,ogden2023examining,vreugdenhil2023reasoning,sokol2025artificial}. To systematically capture these challenges, we decompose medical complexity into \textbf{seven} primary dimensions reflecting both perceptual and reasoning aspects.

From the visual perspective, complexity stems from ambiguity and variability in medical images. We identify several major factors that contribute to perceptual complexity.
\textbf{(1) Small-object detection:} Focuses on identifying small or faint object that are difficult to perceive. \textbf{(2) Fine-detail discrimination:} Addresses distinguishing visually similar findings that may differ in clinical meaning.  \textbf{(3) Spatial understanding:} Involves the integration and alignment of multimodal information to maintain spatial consistency.  

From the reasoning perspective, the following factors play a central role in shaping complexity during higher-order inference and decision-making under uncertainty
\textbf{(4) Temporal prediction:} Concerns reasoning about disease progression and prognosis over time.  
\textbf{(5) Causal reasoning:} Involves connecting symptoms, findings, and outcomes through multi-step causal chains.  
\textbf{(6) Long-tail generalization:} Captures decision-making when rare cases have very few samples compared with common cases.
\textbf{(7) Multi-source integration:} Focuses on extracting key diagnostic cues from multiple co-existing abnormalities within complex cases.
Together, these dimensions reflect the dual nature of medical complexity: perceptual and reasoning, and serve as the conceptual backbone for data selection, question formulation, and evaluation in \method.

\subsection{Benchmark Curation Pipeline}
We begin with collecting medical images that correspond to the seven medical complexity dimensions defined in Sec.~\hyperref[sec:definition]{2.1}, ensuring clinical grounding and diversity.
Next, we design a set of carefully constructed question templates to guarantee strong visual dependence and reasoning complexity in the questions.
These templates are expanded through both automated and manual generation to produce multiple-choice (MCQ) and open-ended questions. 
Extensive human and model-based filtering is then performed to remove low-quality or trivial cases. 
Finally, consistency and correctness checks are conducted by multiple annotators to ensure overall quality assurance.

\vspace{0.5em}

\noindent\textbf{Dimension-guided Data Collection.} 
Building upon the complex clinical reasoning criteria defined in Sec.~\hyperref[sec:definition]{2.1}, we collect data from a wide range of established clinical case reports and research articles published in well-recognized biomedical journals such as JMCR and NEJM. We obtain medical images together with human-annotated captions and relevant metadata. Based on this data, we construct seven question categories: \textit{small-object detection, fine-detail discrimination, spatial understanding, temporal prediction, causal reasoning, long-tail generalization, and multi-source integration} — which together comprehensively cover the defined complexity dimensions. 
The detailed correspondence between complexity dimensions and question categories are provided in the Supplementary Materials.


\noindent\textbf{Question Generation.} 
\label{sec:qg}
We ask annotators with medical backgrounds to design 10–20 templates for each question category.
Each template is designed to maintain strong visual grounding, assess the model’s ability relevant to its medical complexity dimension, and encourage multi-step diagnostic inference beyond direct visual recognition.
 We employ GPT-5\allowbreak -mini\allowbreak -2025\allowbreak -08\allowbreak -07 to assist in automatic question generation: for each collected image, the model selects a suitable template and extracts the correct answer from the corresponding human-annotated caption.  
This process ensures that all generated VQA items are correct, diverse, and focused on the intended reasoning type.

  
\noindent\textbf{Distractor Annotation.} To construct high-quality distractors for multiple-choice questions, we adopt a human-in-the-loop framework for distractor construction. We leverage three large multimodal models: %
GPT-5\allowbreak -Mini\allowbreak -2025\allowbreak -08\allowbreak -07, 
Qwen3\allowbreak -VL\allowbreak -Plus\allowbreak -2025\allowbreak -09\allowbreak -23, and 
Claude\allowbreak -Sonnet\allowbreak -4\allowbreak -20250514. Each generates four distractor candidates for every question. 
This produces a pool of twelve candidates, from which three human annotators with medical backgrounds select four final distractors. 
The final four distractors must satisfy four criteria: 
(1) Sufficient difficulty, ensuring that the question remains challenging for well-prepared test-takers; 
(2) Correctness exclusion, guaranteeing that distractors are always false and not semantically overlapping with the correct answer; 
(3) visual dependence, requiring that items must rely on the interpretation of visual information rather than text alone; and
(4) Clinical plausibility, ensuring that each option remains realistic and contextually appropriate in medical scenarios.


\noindent\textbf{Data Filtering.}  
We apply model-based filtering and human check to select high-quality and challenging VQA items for both humans and MLLMs.


\noindent\textbf{(1) Human Filtering.}  
Before \hyperref[sec:qg]{Question Generation}, healthcare practitioners manually review all collected images to remove those with insufficient caption, or a mismatch with the target evaluation dimension. This step ensures that the retained images are high-quality and aligned with the specific capabilities to be evaluated in MLLMs.


\noindent\textbf{(2) Model Filtering.} After~\mbox{\hyperref[sec:qg]{Question Generation}}, we evaluate each item using three multimodal models: Lingshu\allowbreak-7B, Qwen2.5\allowbreak-VL\allowbreak-7B, and Llava\allowbreak-Med\allowbreak-v1.5\allowbreak-Mistral\allowbreak-7B. Questions that all three models answer correctly are excluded, ensuring appropriate difficulty for MLLMs.


\noindent\textbf{Quality Assurance.}
We implemented multi-stage quality verification procedures to safeguard data integrity and medical validity. General practitioners were incorporated into the screening process, with a specific focus on filtering LLM-generated data—resulting in the ultimate elimination of 8\% of questionable content from the original synthetic dataset. During the human-in-the-loop filtering phase, two annotators jointly conducted manual review, while an independent auditor validated the consistency of their judgments; questions failing to reach consensus were excised. Subsequent to model-driven filtering, four annotators verified that each remaining question possessed a single, unambiguous correct answer, and any entry with other potentially valid options was removed. Finally, all questions underwent a comprehensive review by licensed physicians to confirm their medical accuracy.


\subsection{Benchmark Statistics}
In total, \method~includes 20,653 questions, comprising 16,655 multiple-choice questions (MCQs), each with five candidate options, and 3,998 open-ended questions.

\begin{figure*}[t]
    \centering
    \resizebox{\linewidth}{!}{
    \includegraphics[width=\linewidth]{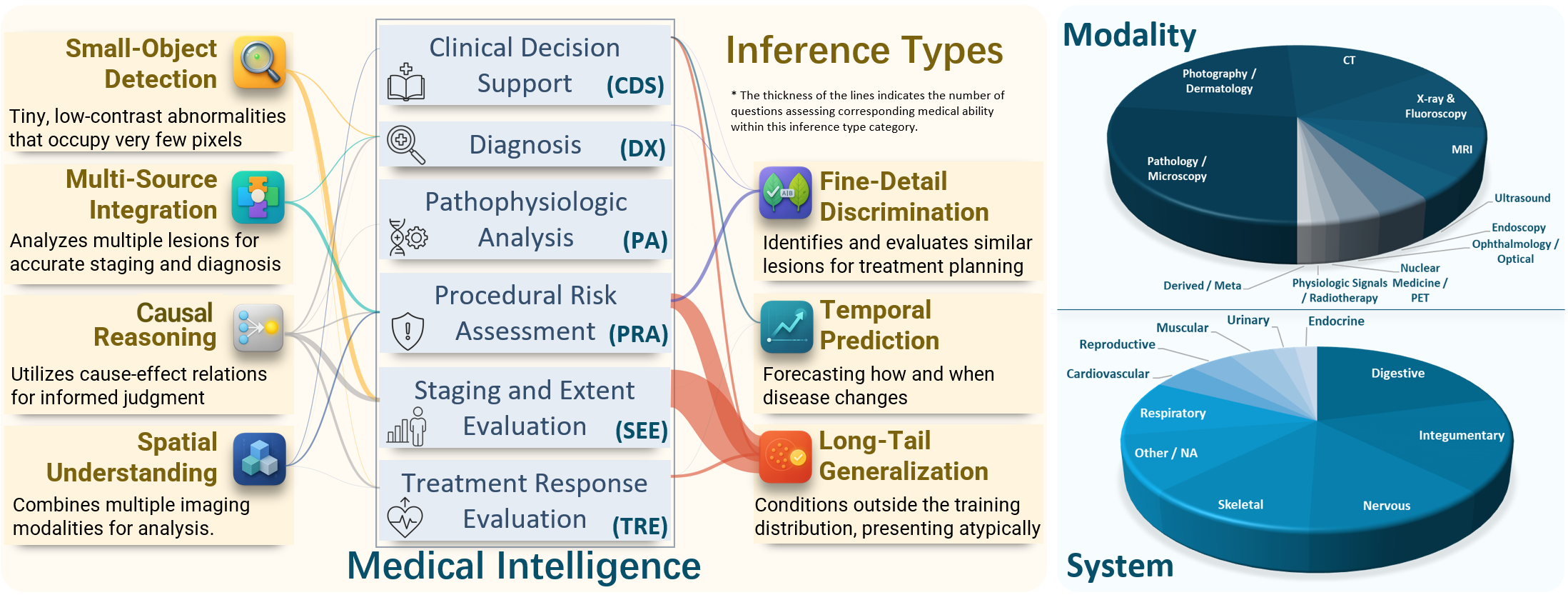}
    }
    \vspace{-7mm}
    \caption{Benchmark statistics. The left panel displays the inference types across seven questions in the benchmark and their corresponding quantitative relationships with medical ability. The right side shows the modalities of the benchmark images and the body systems involved.}
    \vspace{-5mm}
    \label{fig:stat}
\end{figure*}

All questions are systematically organized into seven primary categories introduced in Sec.\hyperref[sec:definition]{3.2}, each reflecting a distinct aspect of clinical reasoning. Within each category, five additional subtypes are introduced to provide a more fine-grained classification of question types.
Furthermore, comprehensive domain coverage is achieved in our benchmark, spanning \textbf{11} body systems and \textbf{12} imaging modalities, following authoritative medical ontologies.
The body system taxonomy follows the standard classification proposed by Liachovitzky~\cite{liachovitzky2015human}.
The imaging modality taxonomy follows the \textit{Digital Imaging and Communications in Medicine (DICOM)} and the \textit{RSNA RadLex Playbook}~\cite{DICOM,RSNA_RadLex_Playbook_2018}, 
and is further extended beyond radiology to include modalities such as pathology, ophthalmology, endoscopy. The complete classification standards are provided in Supplementary Materials. Figure~\ref{fig:stat} summarizes all benchmark statistics.

\subsection{Comprehensive Evaluation Framework}
We develop a multi-phase evaluation framework that measures model performance across both structured and open-ended tasks \cite{liu2025human}. For multiple-choice questions (MCQs), the evaluation is based on the correctness. For open-ended questions, we use a criterion-based scoring scheme that evaluates both the reasoning process and the final outcome.
Each open-ended response is scored based on four predefined dimensions.
The final score \( S \) for open-ended questions is computed as a weighted sum of four dimension scores:
\[
S = \frac{\sum_{i \in \{\text{cons},\,\text{coh},\,\text{vis},\,\text{gt}\}} w_i \, s_i}{\sum_{i \in \{\text{cons},\,\text{coh},\,\text{vis},\,\text{gt}\}} w_i},
\]
where \( s_i \) denotes the score for each dimension \( i \), and the weights are assigned as 
\( w_{\text{cons}} = 1 \), \( w_{\text{coh}} = 1 \), \( w_{\text{vis}} = 4 \), and \( w_{\text{gt}} = 4 \).
Each dimension measures a distinct aspect of response quality:  
\vspace{0.3em}
\begin{itemize} 
    \item \textbf{Consistency} — Clarity and internal agreement
    \item \textbf{Coherence} — Causal connection across reasoning steps
    \item \textbf{Visual accuracy} —  Accuracy in recognizing and describing visual features in the image
    \item \textbf{Ground-truth correctness} — Alignment of the final answer with ground truth
\end{itemize}
We employ DeepSeek-V3.2-Exp as an independent evaluator under a controlled protocol to ensure scoring quality and minimize bias toward the tested models.
The scoring prompt is provided in Supplementary Material. 
In Sec.\hyperref[sec:experiments]{3}, all open-ended scores are normalized to a 0--100 scale for consistent comparison.

\section{Experiments}
\label{sec:experiments}

\subsection{Experimental Setup}

\begin{table*}[h]
    \centering
        \vspace{-5mm}
        \caption{Scores of MLLMs on the benchmark. Seven inference types are Small-Object Detection (SOD), Fine-Detail Discrimination (FDD), Spatial Understanding (SU), Temporal Prediction (TP), Causal Reasoning (CR), Long-Tail Generalization (LTG), and Multi-Source Integration (MSI). Four metrics for open-ended questions are Consistency (Con), Coherence (Coh), Visual accuracy (VA), and Ground-truth correctness (GT). \textbf{Bold} indicates the best. \underline{underline} indicates the second place. Closed-source models and open-source models of different sizes are ranked separately.}\label{tbl:full_score}
\vspace{-3mm}
\scriptsize
\resizebox{\textwidth}{!}{
        \setlength{\tabcolsep}{1mm}{

\begin{tabular}{ll | cccccccc|ccccc}
\toprule
\makecell{Model} & 
\makecell{Year} & 
\makecell{SOD\\\includegraphics[trim={0 150px 0 150px},width=22px]{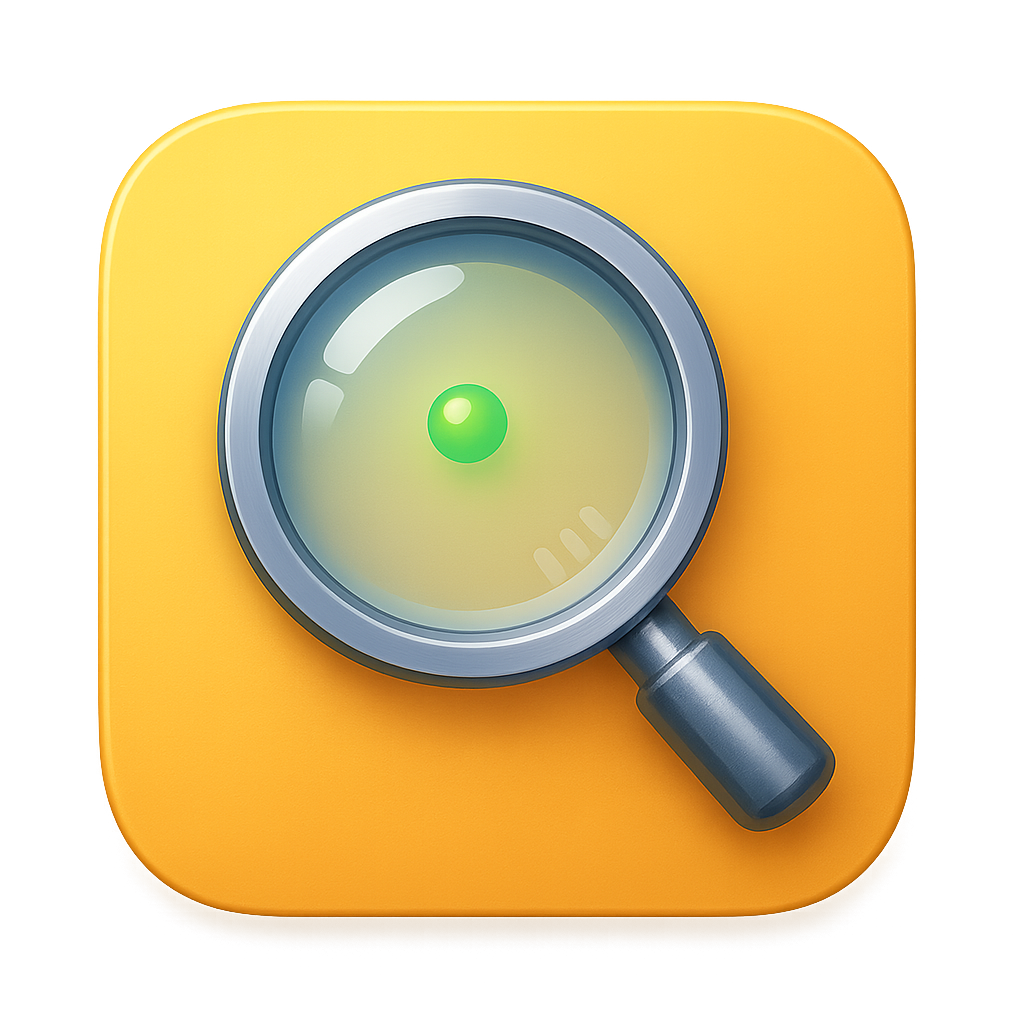}} &
\makecell{FDD\\\includegraphics[trim={0 150px 0 150px},width=22px]{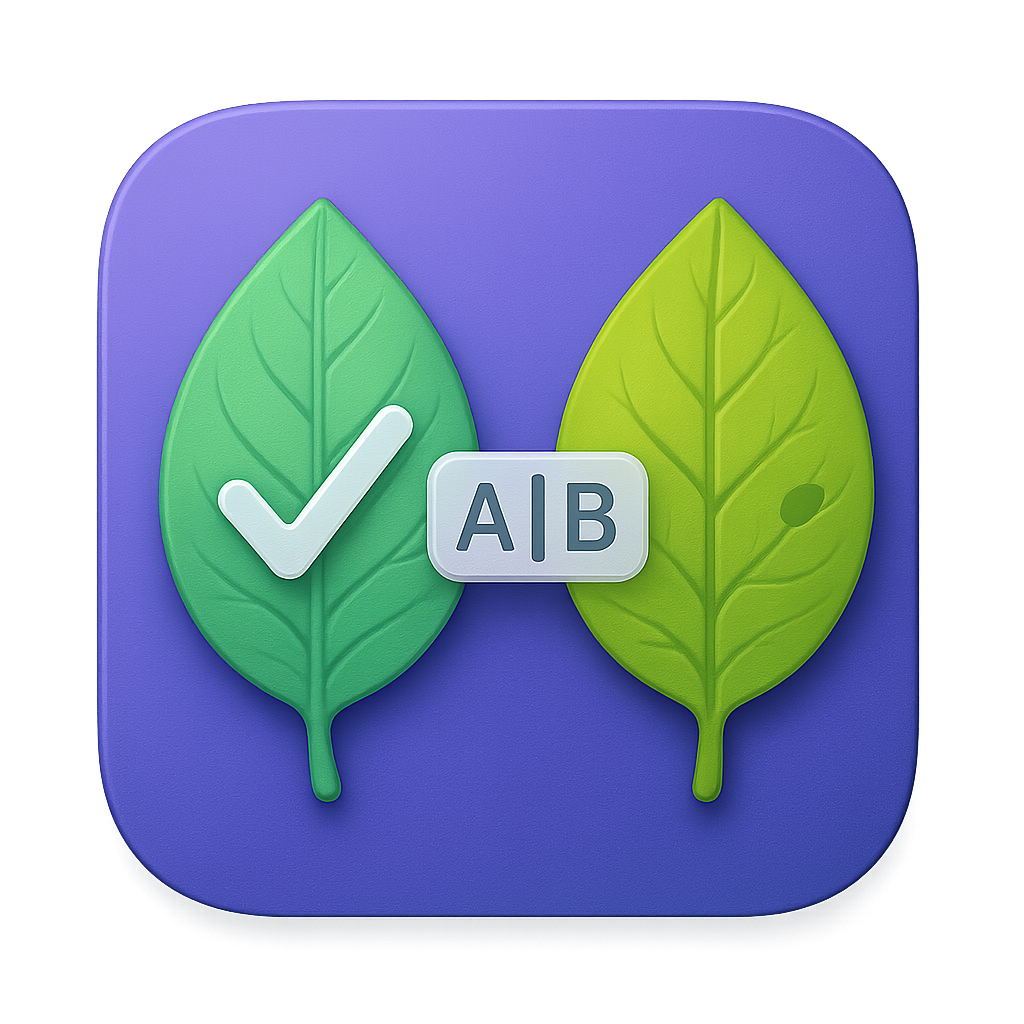}} &
\makecell{SU\\\includegraphics[trim={0 150px 0 150px},width=22px]{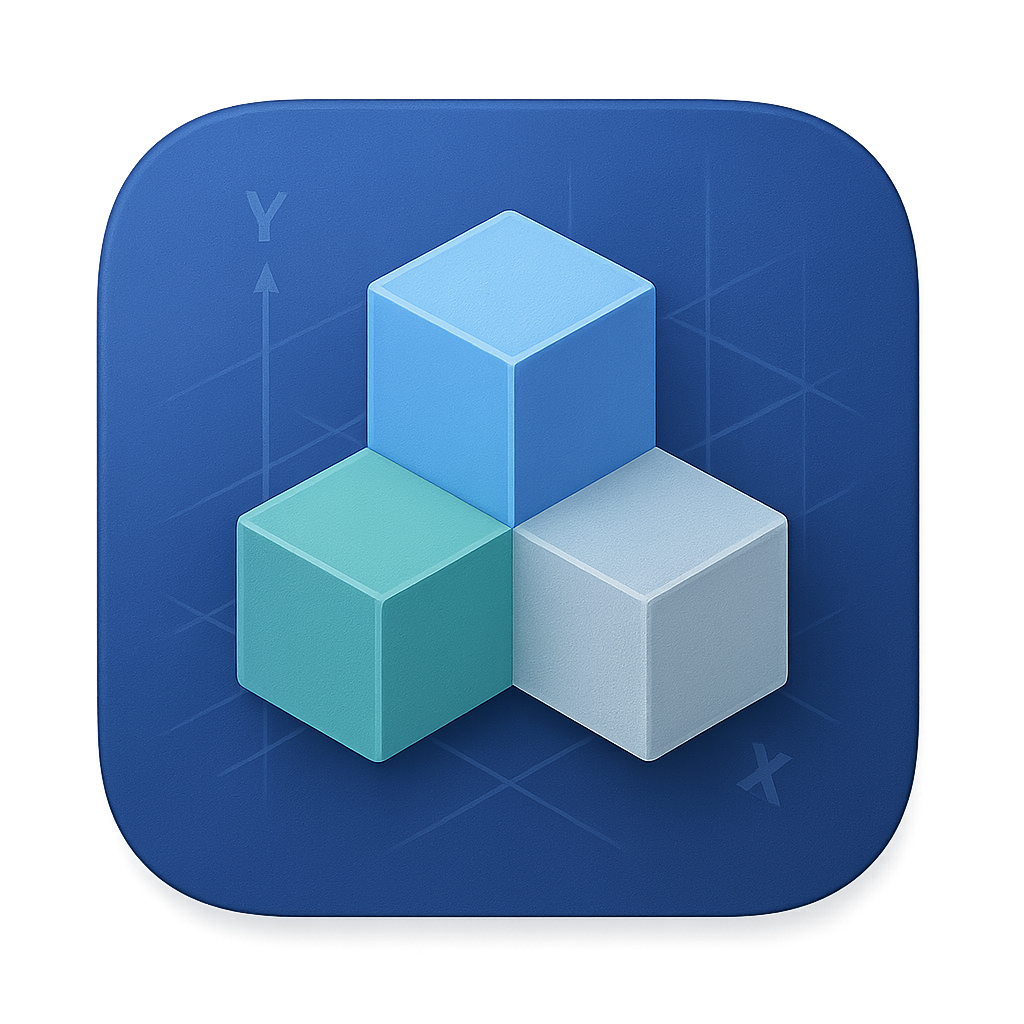}} &
\makecell{TP\\\includegraphics[trim={0 150px 0 150px},width=22px]{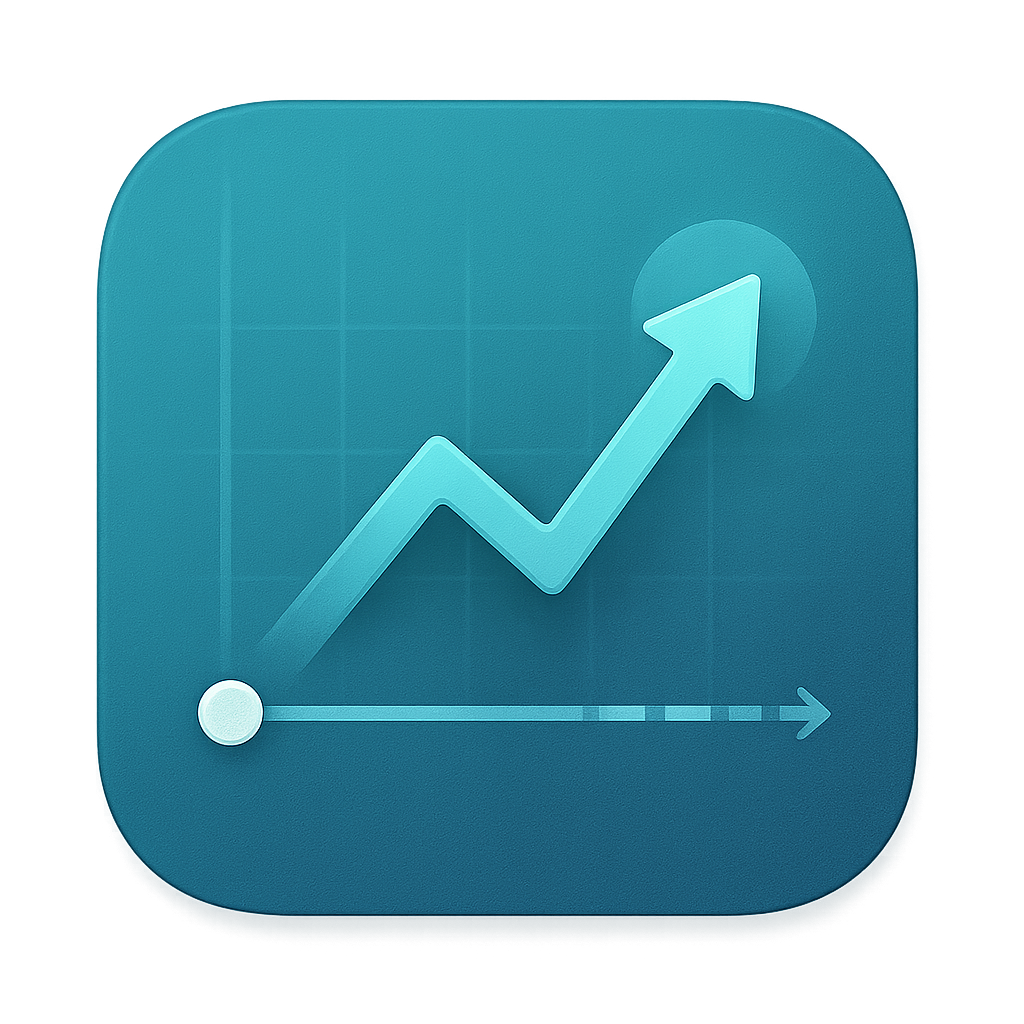}} & 
\makecell{CR\\\includegraphics[trim={0 150px 0 150px},width=22px]{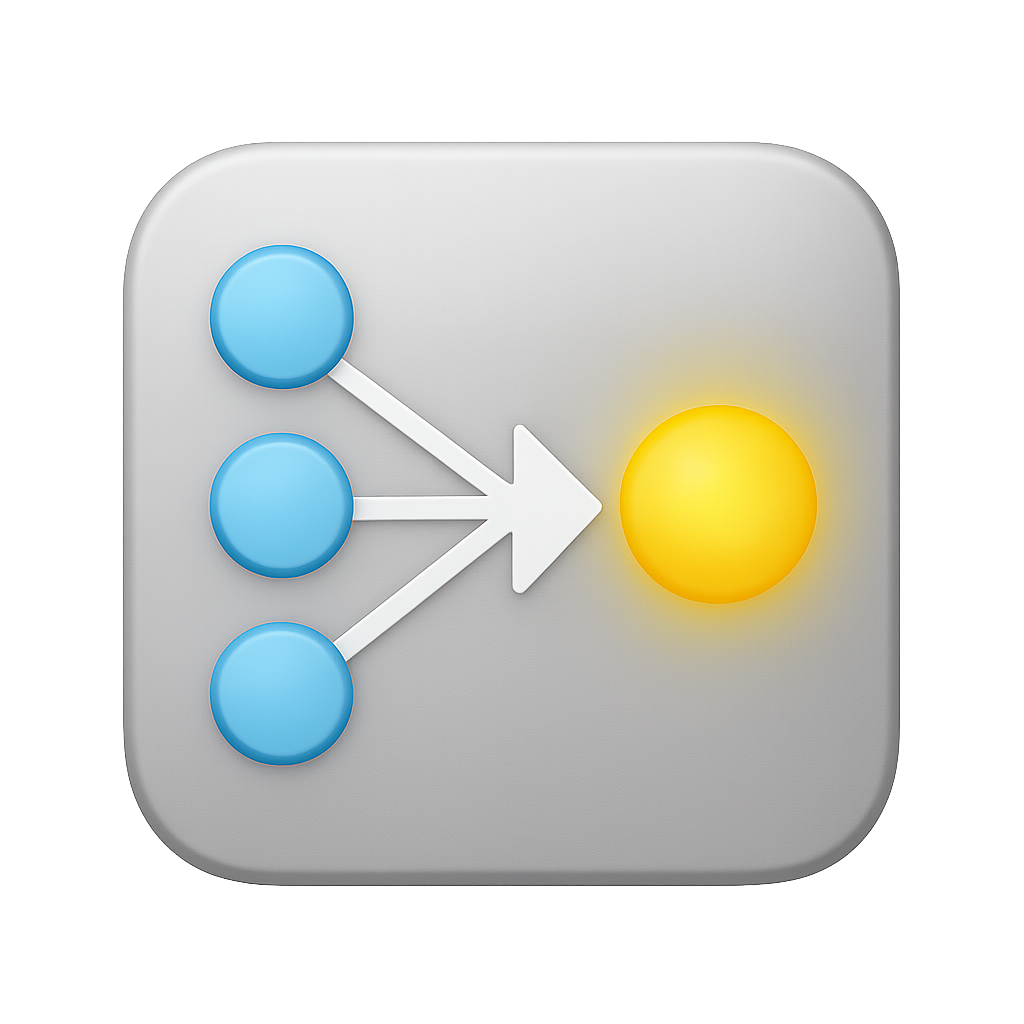}} & 

\makecell{LTG\\\includegraphics[trim={0 150px 0 150px},width=22px]{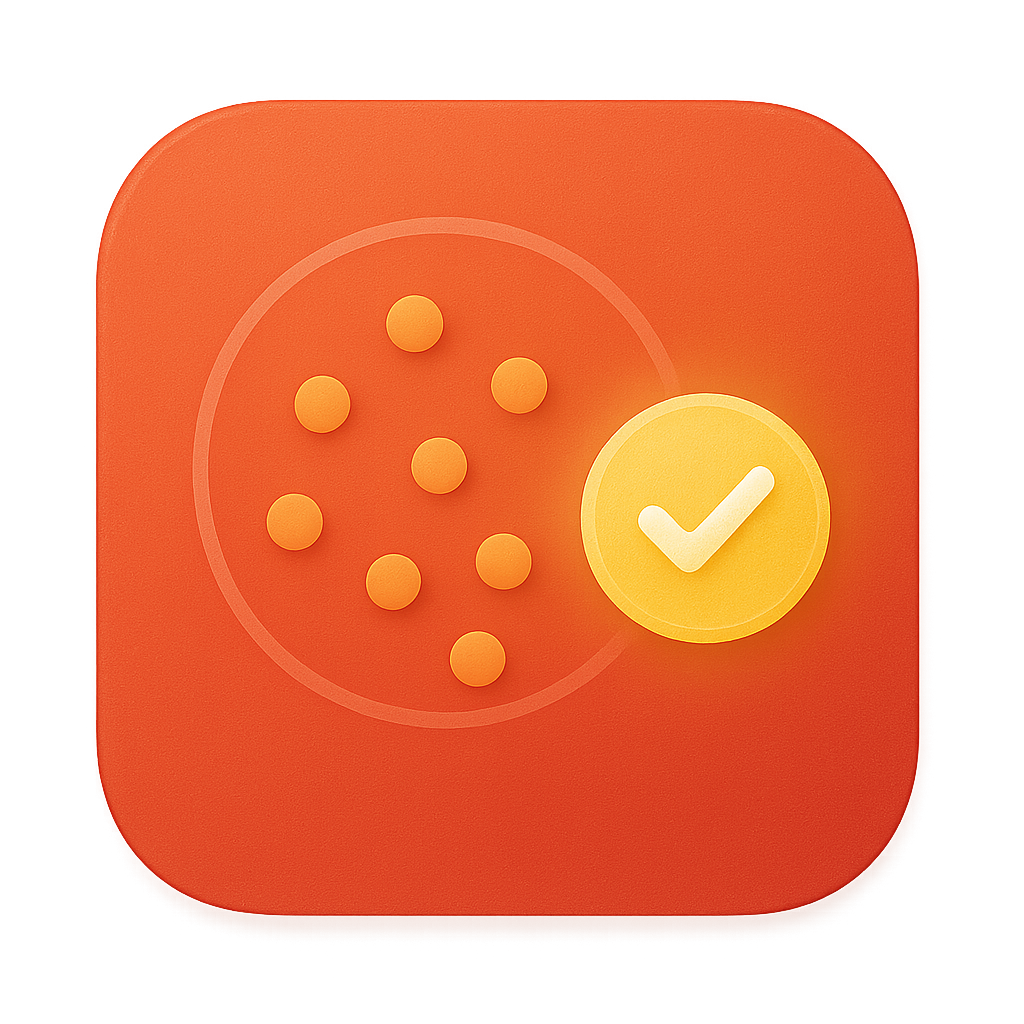}} &
\makecell{MSI\\\includegraphics[trim={0 150px 0 150px},width=22px]{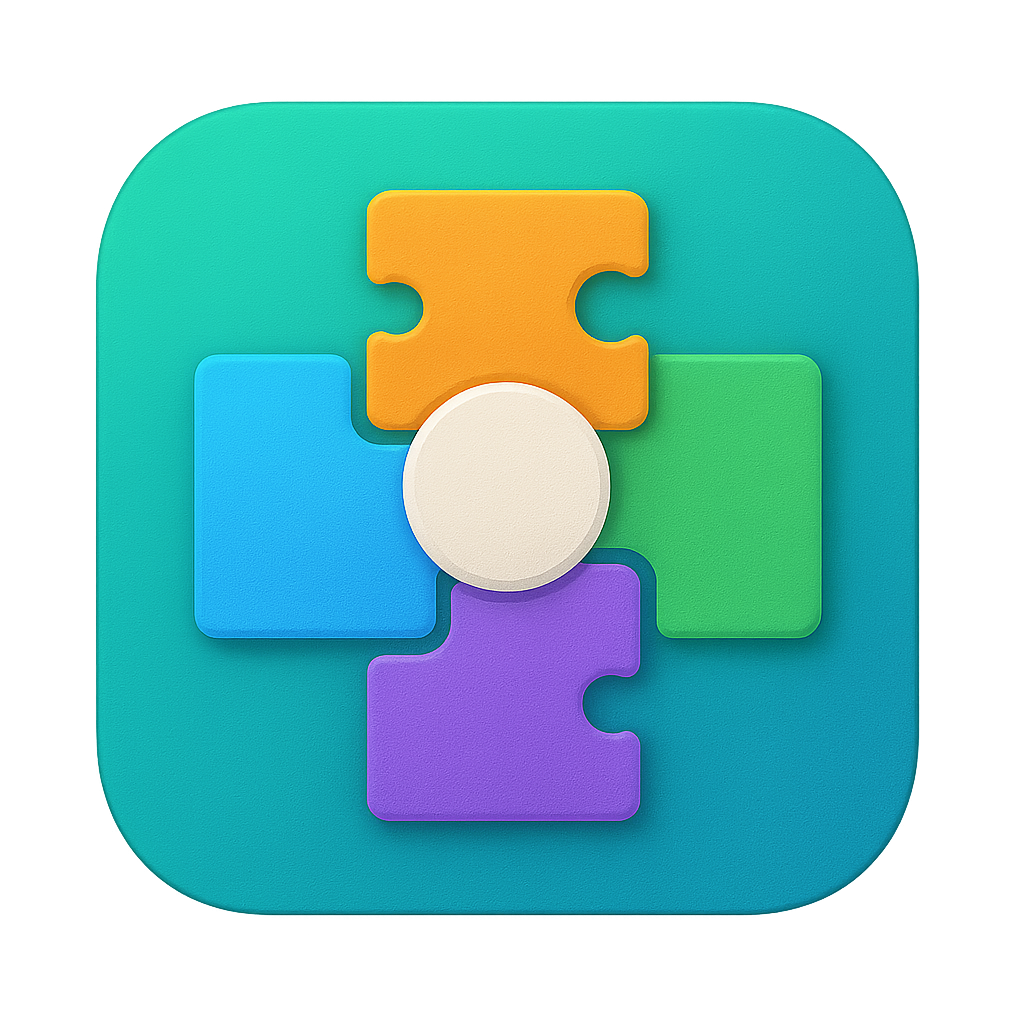}} &

\makecell{All Score\\\includegraphics[trim={0 150px 0 150px},width=22px]{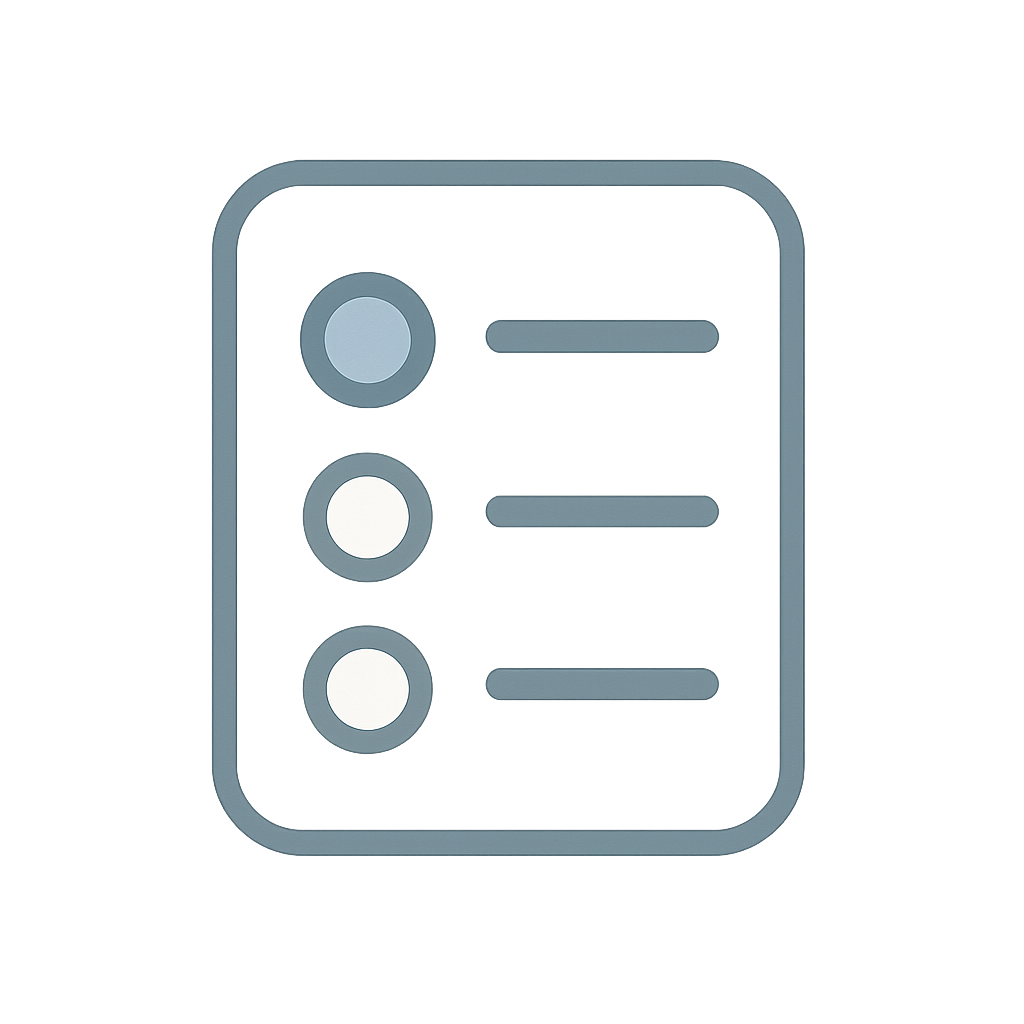}} & \makecell{\makecell{Con}} & \makecell{\makecell{Coh}} & \makecell{VA} & \makecell{GT} & \makecell{All Score\\\includegraphics[trim={0 150px 0 150px},width=22px]{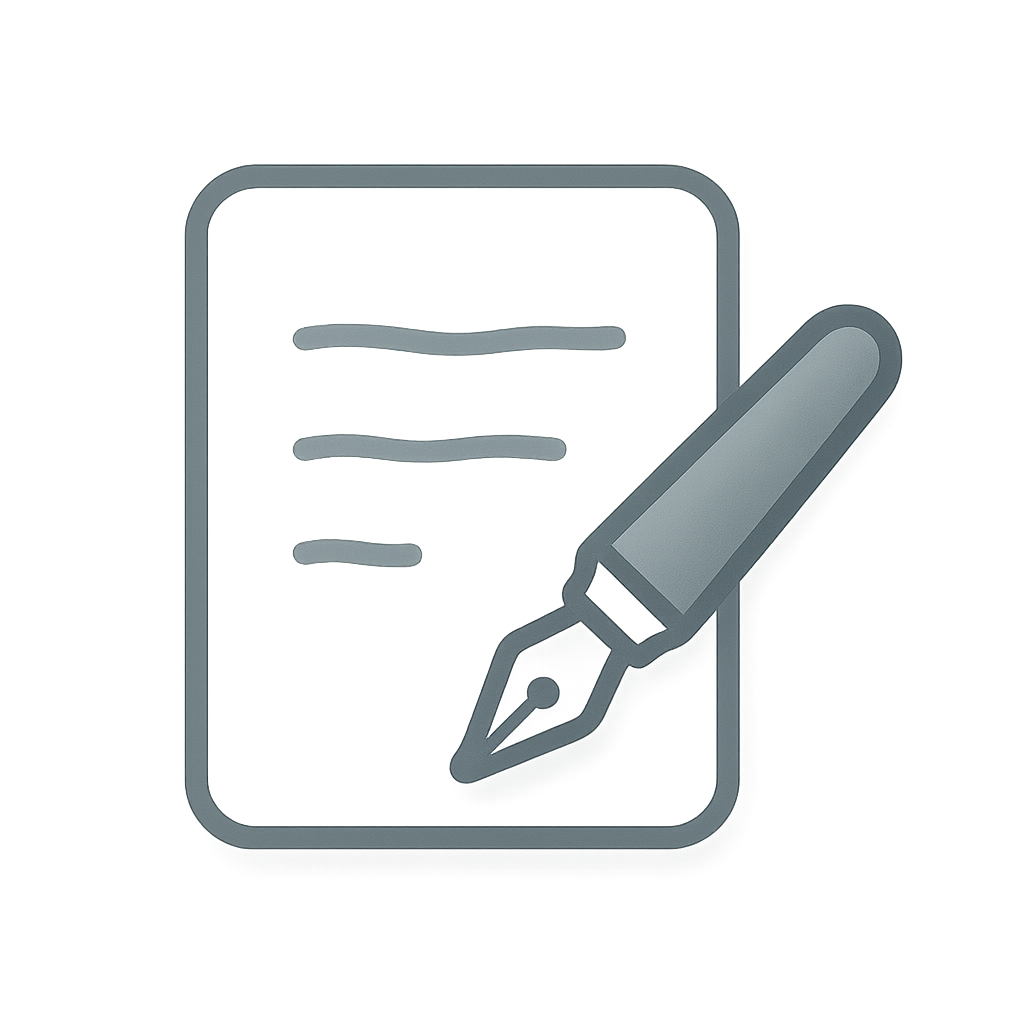}} \\
\midrule
\multicolumn{14}{c}{Closed-source Models}\\
\hline
\rowcolor{gray!15}
GPT-5 \cite{openai_gpt5_systemcard_2025} & 2025 & \textbf{66.08} & \textbf{71.45} & \textbf{62.06} & \textbf{58.33} & \textbf{60.30} & \textbf{55.19} & \textbf{69.00} & \textbf{57.81} & \underline{97.77} & \underline{88.86} & \textbf{40.45} & \textbf{34.65} & \textbf{48.70}\\
Gemini-2.5-Pro \cite{comanici2025gemini25} & 2025& \underline{58.75} & \underline{68.07} & \underline{56.70} & \underline{52.08} & \underline{53.54} & \underline{46.42} & \underline{64.42} & \underline{49.87} & \textbf{98.11} & \textbf{89.36} & \underline{35.77} & \underline{32.07} & \underline{45.98} \\ \hline
\multicolumn{14}{c}{Open-source Models \textbf{1B\textasciitilde
 10B}}\\
\hline
\rowcolor{gray!15}
Medgemma-4B  \cite{sellergren2025medgemma} & 2025 & 16.13 & 17.72 & 13.12 & 14.58 & 17.64 & 14.00 & 23.45 & 14.90 & 86.98 & 57.17 & 26.65 & 17.57 & 32.10\\
Lingshu-7B \cite{lasateam2025lingshu} & 2025& \text{32.84} & \text{47.12} & 31.17 & \underline{38.99} & \text{31.53} & \text{23.86} & \text{39.62} & \text{27.26} & \textbf{96.19} & \textbf{73.96} & \underline{33.47} & \underline{26.26} & \underline{40.91}\\
\rowcolor{gray!15}
Gemma3-4B  \cite{team2025gemma} & 2025 & 31.57 & 38.68 & \text{31.59} & 25.00 & 28.10 & 23.74 & 35.04 & 25.98 & 90.05 & 64.26 & 18.53 & 13.14 & 28.10\\
Qwen2.5-VL-7B \cite{Qwen2.5-VL} & 2025& \underline{37.83} & \underline{48.10} & \underline{33.29} & \text{32.74} & \underline{35.22} & \underline{28.15} & \underline{43.13} & \underline{31.06} & \text{92.35} & 65.83 & 26.21 & 19.15 & 33.96\\
\rowcolor{gray!15}
InternVL3.5-8B \cite{wang2025internvl3_5} & 2025& 29.52 & 36.01 & 25.95 & 30.95 & 29.71 & 21.53 & 31.00 & 24.17 & \underline{95.20} & \text{71.10} & \textbf{35.38} & \textbf{26.65} & \textbf{41.44}\\
Qwen3-VL-8B \cite{qwen3technicalreport} & 2025 & \textbf{46.63} & \textbf{53.87} & \textbf{45.84} & \textbf{42.86} & \textbf{43.50} & \textbf{34.48} & \textbf{53.64} & \textbf{38.18} & 88.46 & \underline{72.99} & \text{28.72} & \text{23.54} & \text{37.05} \\ \hline
\multicolumn{14}{c}{Open-source Models \textbf{10B\textasciitilde
 100B}}\\
\hline
\rowcolor{gray!15}
Medgemma-27B  \cite{sellergren2025medgemma}  & 2025& 37.44 & 47.54 & 37.24 & 29.46 & 30.80 & 25.92 & 36.93 & 28.91 & 89.82 & 72.21 & 20.64 & 17.58 & 31.49\\
Lingshu-32B \cite{lasateam2025lingshu} & 2025& 37.83 & 48.95 & 31.88 & 38.99 & 36.21 & 27.08 & 40.70 & 30.47 & \underline{96.89} & 76.36 & \underline{35.76} & \underline{28.48} & \underline{43.02}\\
\rowcolor{gray!15}
Gemma3-27B  \cite{team2025gemma} & 2025& \text{45.94} & \text{52.74} & \text{45.98} & 37.80 & \text{42.35} & \text{33.62} & 45.28 & \text{37.07} & 93.72 & 76.23 & 25.96 & 19.96 & 35.36\\
Qwen2.5-VL-32B \cite{Qwen2.5-VL} & 2025 & 42.82 & 52.60 & 36.25 & 38.99 & 39.39 & 29.60 & 45.82 & 33.36 & 95.54 & 75.81 & \text{33.55} & \text{26.66} & \text{41.22}\\
\rowcolor{gray!15}
Qwen2.5-VL-72B \cite{Qwen2.5-VL} & 2025 & \textbf{52.10} & \textbf{61.32} & \underline{47.39} & \textbf{51.19} & \underline{46.36} & \underline{38.46} & \textbf{54.18} & \underline{42.17} & \text{96.31} & 75.11 & 33.29 & 25.68 & 40.73\\     
InternVL3.5-38B \cite{wang2025internvl3_5} & 2025 & 42.13 & 48.10 & 37.80 & \text{44.05} & 38.35 & 32.48 & 40.97 & 35.07 & \textbf{97.20} & \text{76.48} & \textbf{36.83} & \textbf{29.03} & \textbf{43.71}\\
\rowcolor{gray!15}
Qwen3-VL-30B-A3B \cite{qwen3technicalreport} & 2025 & 44.57 & 51.20 & 41.18 & 39.88 & 38.14 & 32.86 & \text{47.17} & 35.79 & 94.37 & \textbf{79.84} & 29.73 & 25.15 & 39.37\\
Qwen3-VL-32B \cite{qwen3technicalreport} & 2025 & \underline{49.66} & \underline{60.90} & \textbf{49.22} & \underline{46.43} & \textbf{47.45} & \textbf{41.58} & \underline{53.91} & \textbf{44.28} & 96.00 & \underline{79.31} & 31.09 & 25.65 & 39.37 \\ \hline
\multicolumn{14}{c}{Open-source Models \textbf{Larger than
 100B}}\\
\hline
\rowcolor{gray!15}
InternVL3.5-241B-A28B \cite{wang2025internvl3_5} & 2025 & \underline{55.91} & \underline{65.68} & \underline{52.47} & \underline{54.17} & \underline{48.80} & \underline{42.73} & \underline{56.33} & \underline{46.17} & \underline{96.87} & \underline{76.33} & \textbf{42.74} & \textbf{33.66} & \textbf{47.88}\\
Qwen3-VL-235B-A22B\cite{qwen3technicalreport} & 2025 & \textbf{57.48} & \textbf{66.95} & \textbf{55.99} & \textbf{55.06} & \textbf{53.33} & \textbf{45.86} & \textbf{63.07} & \textbf{49.34} & \textbf{97.10} & \textbf{85.21} & \underline{33.02} & \underline{27.95} & \underline{42.62} \\
\bottomrule

\end{tabular}

        }
}
\vspace{-3mm}

\end{table*} 

\textbf{Evaluated MLLMs.} We evaluate 18 multimodal large language models (MLLMs). For closed-source models, we include GPT-5 \cite{openai_gpt5_systemcard_2025} and Gemini-2.5-Pro \cite{comanici2025gemini25}. For open-source models, we assess Gemma-4B and Gemma-27B \cite{team2025gemma}, Medgemma-4B and Medgemma-27B \cite{sellergren2025medgemma}, Qwen2.5-VL-7B, Qwen2.5-VL-32B, Qwen2.5-VL-72B \cite{Qwen2.5-VL}, InternVL3.5-8B, InternVL3.5-38B and InternVL3.5-241B-A28B \cite{wang2025internvl3_5}, Lingshu-7B and Lingshu-32B \cite{lasateam2025lingshu}, as well as Qwen3-VL-8B, Qwen3-VL-32B, Qwen3-VL-30B-A3B and Qwen3-VL-235B-A22B \cite{qwen3technicalreport}.

\noindent\textbf{Implementation Details.} We deploy the open-source MLLMs using VLLM \citep{kwon2023efficient}, while the closed-source models are accessed through their official APIs. We design prompt templates to constrain the output format of each model and subsequently extract answers using regular-expression matching. The complete prompt templates are provided in the Supplementary Material.

\subsection{Main Results}

\renewcommand{\arraystretch}{1.1}  

Table~\ref{tbl:full_score} shows the performance of two closed-source models and 16 open-source models on \method~benchmark. For multiple-choice questions, closed source models lead by a clear margin. GPT-5 attains the best overall accuracy of 57.81\% and is strongest in every category. Gemini-2.5-Pro ranks second overall at 49.87\% and is second on each subtask. Among open source systems, Qwen3-VL-235B-A22B achieves the highest overall accuracy of 49.34\% and ranks first across all categories. Long-Tail Generalization is consistently the most difficult category, with the top score only 55.19\% and all open-source models below 46\%. Tasks such as Fine-Detail Discrimination and Multi-Source Integration are comparatively easier across models. Model scale helps open source performance but does not close the gap with closed source, leaving a 8.47 point difference between GPT-5 and the best open source result. At the same model size, models fine-tuned on medical datasets, including the Medgemma and Lingshu families, did not demonstrate significant advantages.

For open-ended questions, scores are dominated by the high weights on Visual Accuracy and Ground-truth Correctness. Many models show near-ceiling Consistency and strong Coherence while lagging on visual grounding, for example, Qwen3-VL-30b-A3B reaches the highest Coherence within the open source set under 100B at 79.84 yet remains below thirty on the two vision-weighted dimensions. These results indicate that fluent and internally consistent reasoning is relatively mature, while extracting the right visual evidence and converging to the correct answer remain the principal bottlenecks. The advantage of closed source models persists but the gap to the best open-source model is clearly reduced, at 0.82 points on the overall score.

\begin{figure*}[t]
    \centering
    \resizebox{\linewidth}{!}{
    \includegraphics[width=\linewidth]{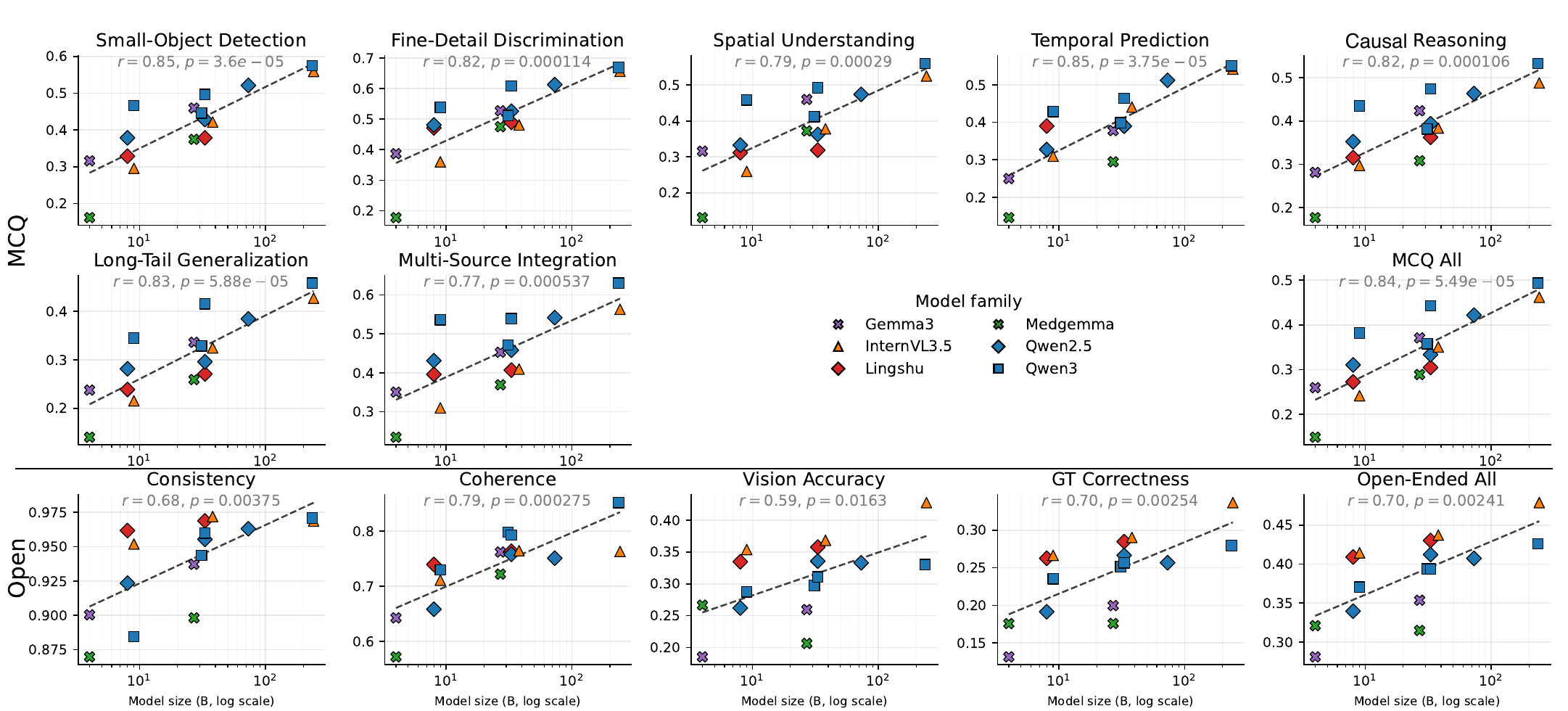}
    }
    \vspace{-5mm}
    \caption[Correlation between model size and performance]{Correlation between model size and performance in different metrics.}
    \vspace{-3mm}
    \label{fig:metric-corr}
\end{figure*}

Figure~\ref{fig:metric-corr} analyzes the correlation between the model's performance across different metrics and its own size. In multiple-choice questions, scaling consistently improves accuracy across all categories, showing that larger models benefit from greater capacity for pattern recognition, multimodal integration, and clinical knowledge retrieval. This trend indicates that most perceptual abilities and knowledge accuracy in structured diagnostic settings can be enhanced through model expansion. However, in open-ended reasoning, the advantage of scale is concentrated on linguistic quality rather than visual understanding. Bigger models produce more coherent and internally consistent explanations, but their gains in visual grounding and factual correctness are much weaker, suggesting that increasing parameter count alone does not substantially improve the ability to extract and apply visual evidence. It reveals that open-ended reasoning requires advances beyond mere size enlargement.

\begin{table}[t!]
    \centering
        \caption{Scores of MLLMs on the benchmark by medical intelligence in Figure~\ref{fig:stat}. Medical intelligence includes clinical decision support (CDS), diagnosis (DX), psychophysiologic analysis (PA), procedural risk assessment (PRA), staging and extent evaluation (SEE), and treatment response evaluation (TRE)}\label{tbl:full_score_app}
\vspace{-3mm}
\scriptsize
\resizebox{\linewidth}{!}{
        \setlength{\tabcolsep}{0.1mm}{

\begin{tabular}{l | cccccc}
\toprule
\makecell{Model} & \makecell{CDS\\\includegraphics[trim={0 150px 0 150px},width=22px]{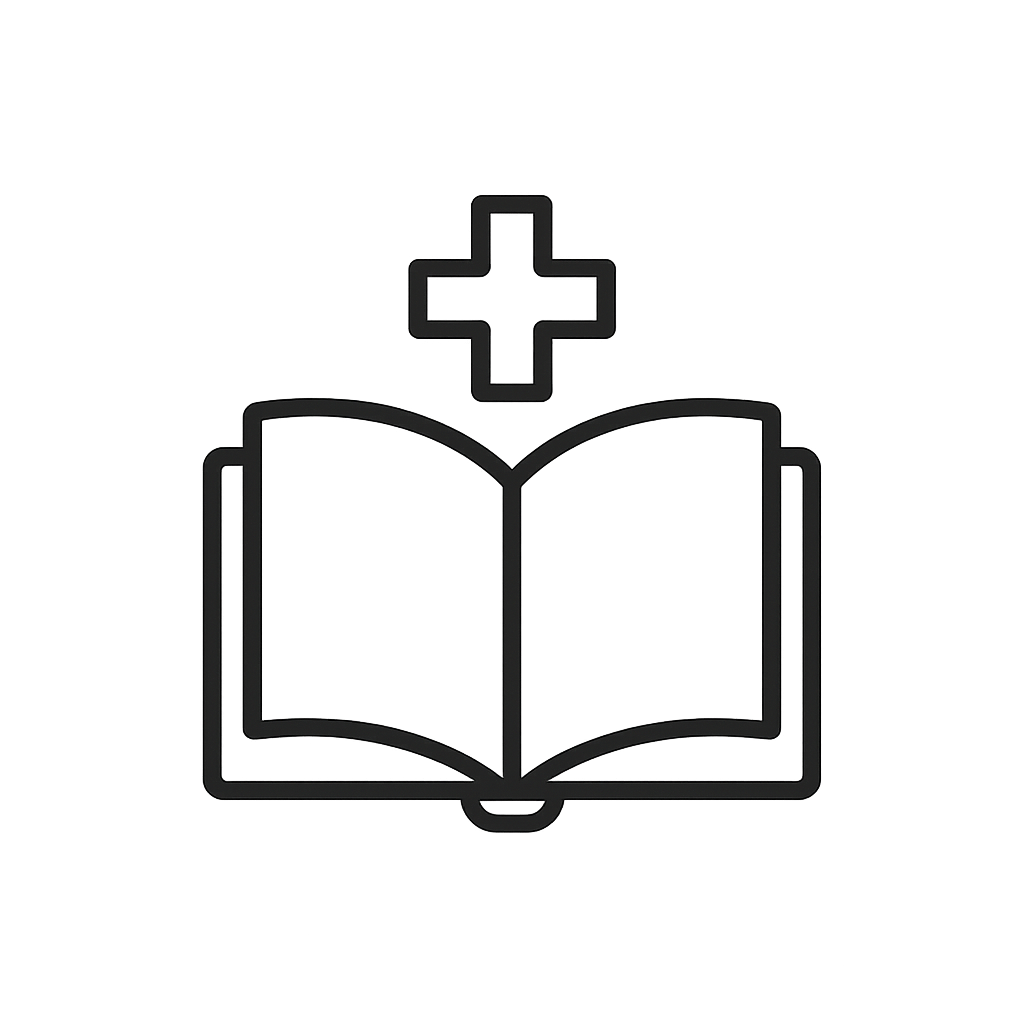}} & \makecell{DX\\\includegraphics[trim={0 150px 0 150px},width=22px]{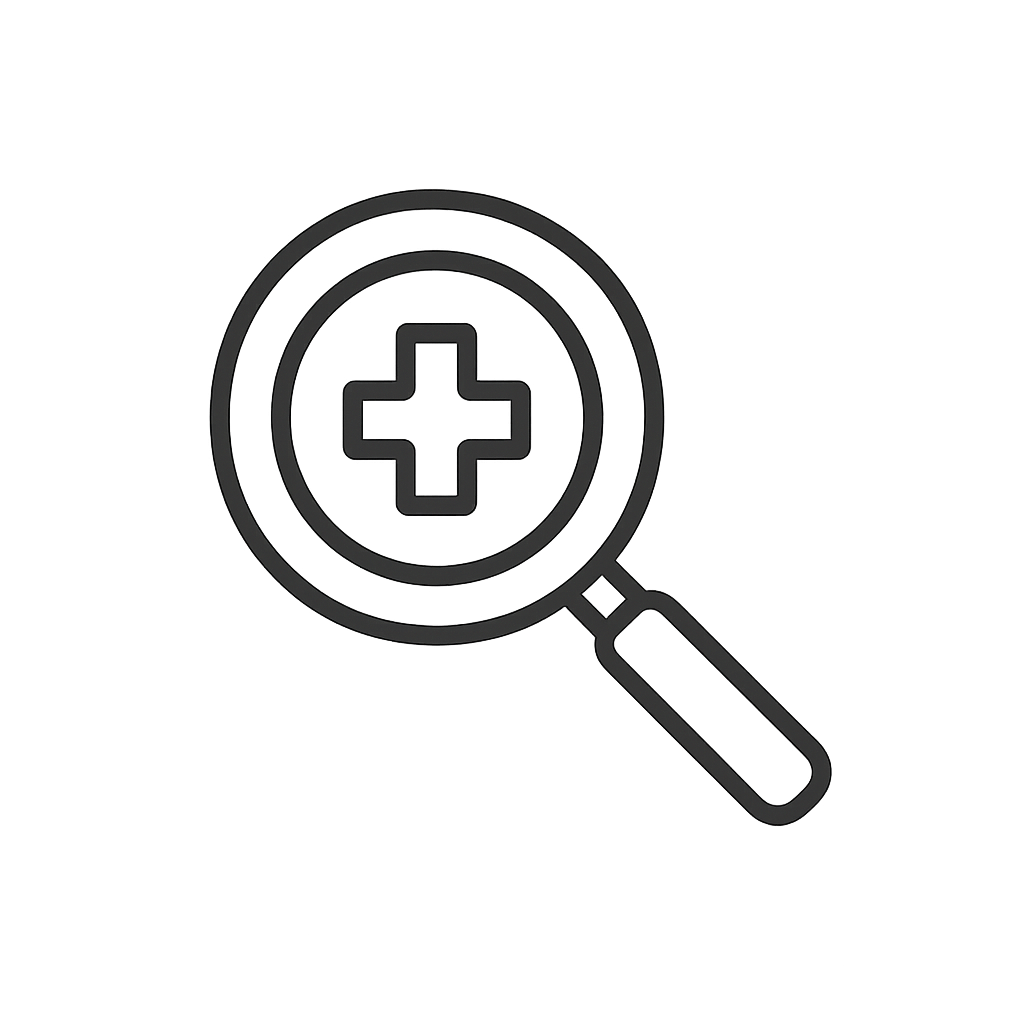}} & \makecell{PA\\\includegraphics[trim={0 150px 0 150px},width=22px]{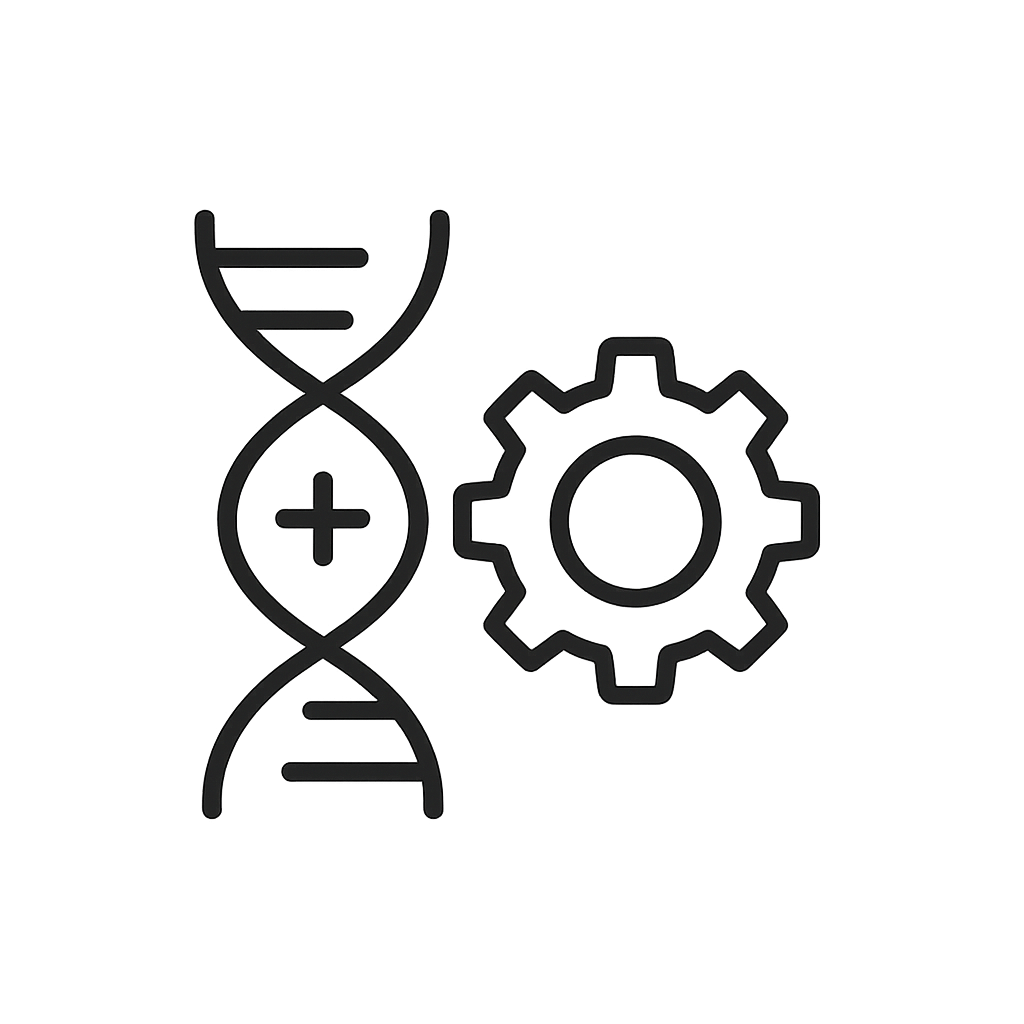}} & \makecell{PRA\\\includegraphics[trim={0 150px 0 150px},width=22px]{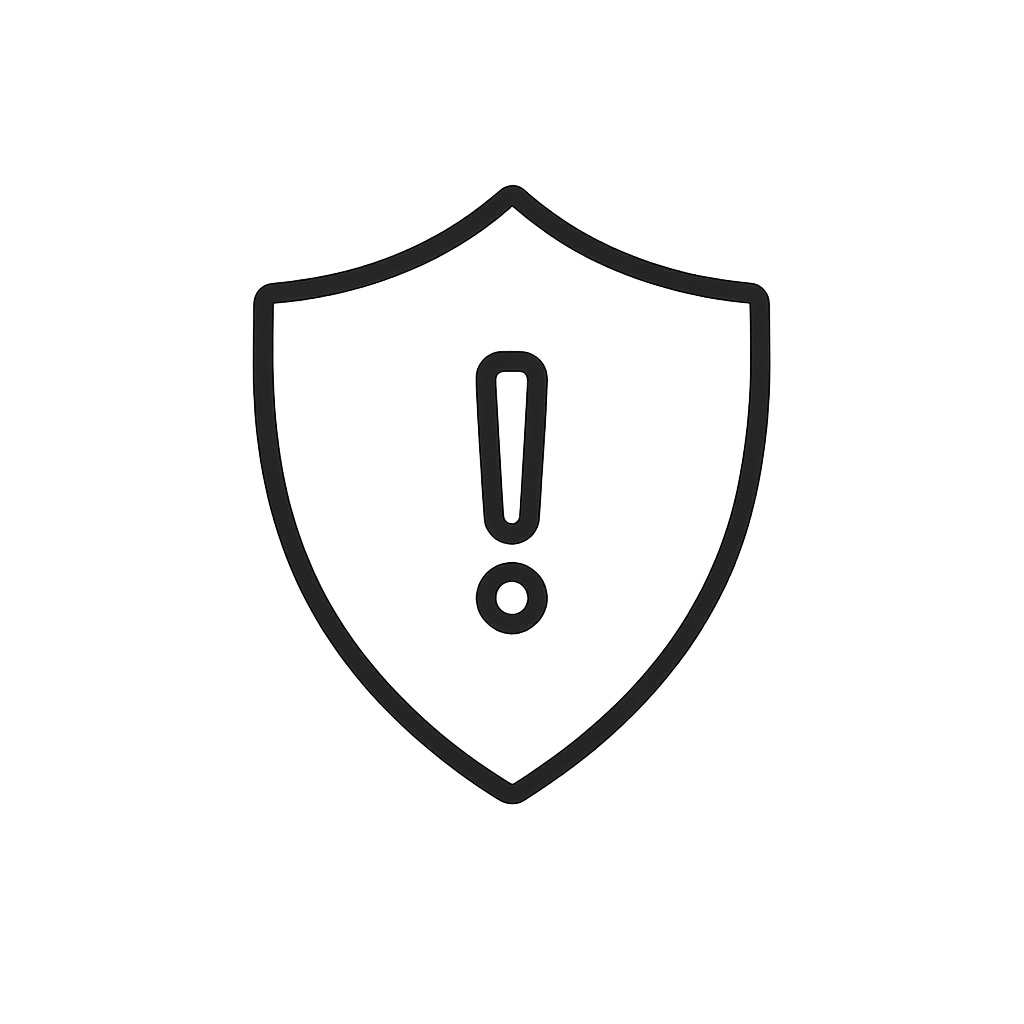}} & \makecell{SEE\\\includegraphics[trim={0 150px 0 150px},width=22px]{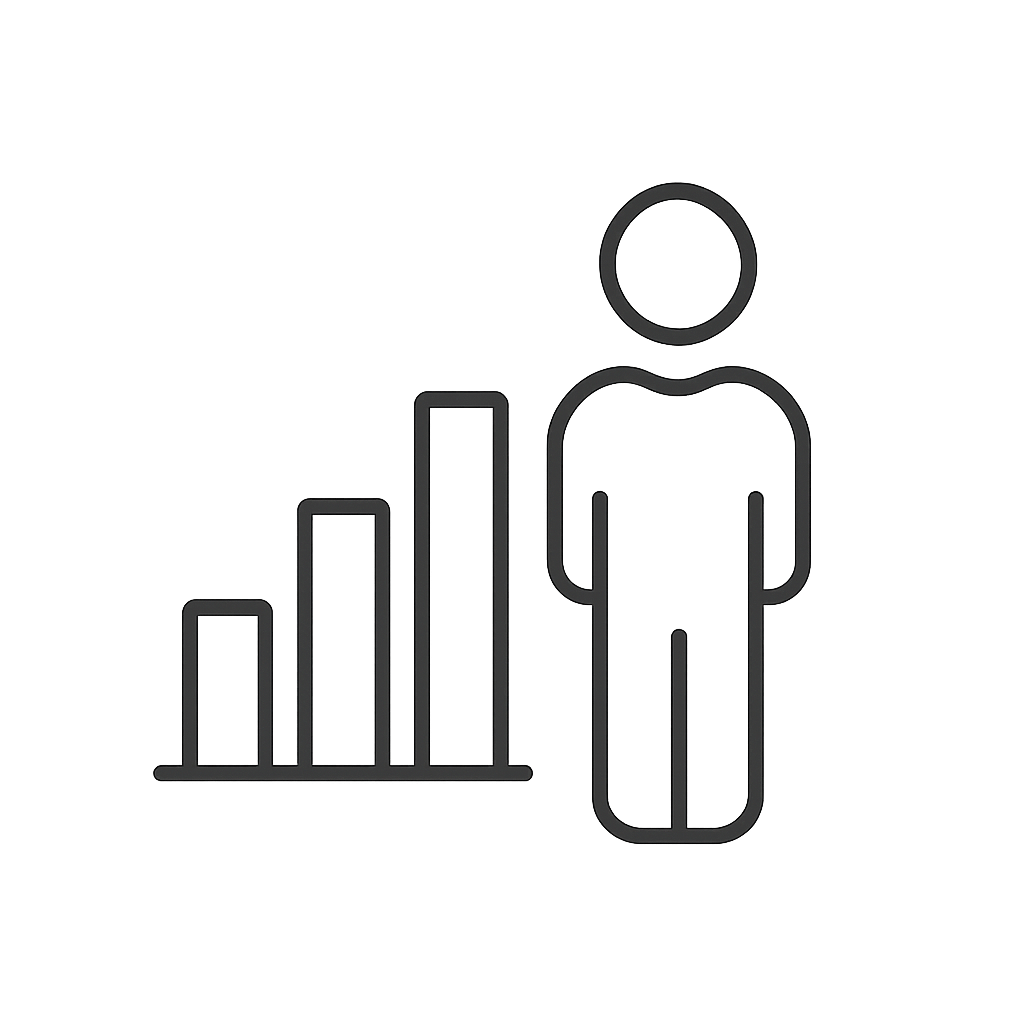}} & \makecell{TRE\\\includegraphics[trim={0 150px 0 150px},width=22px]{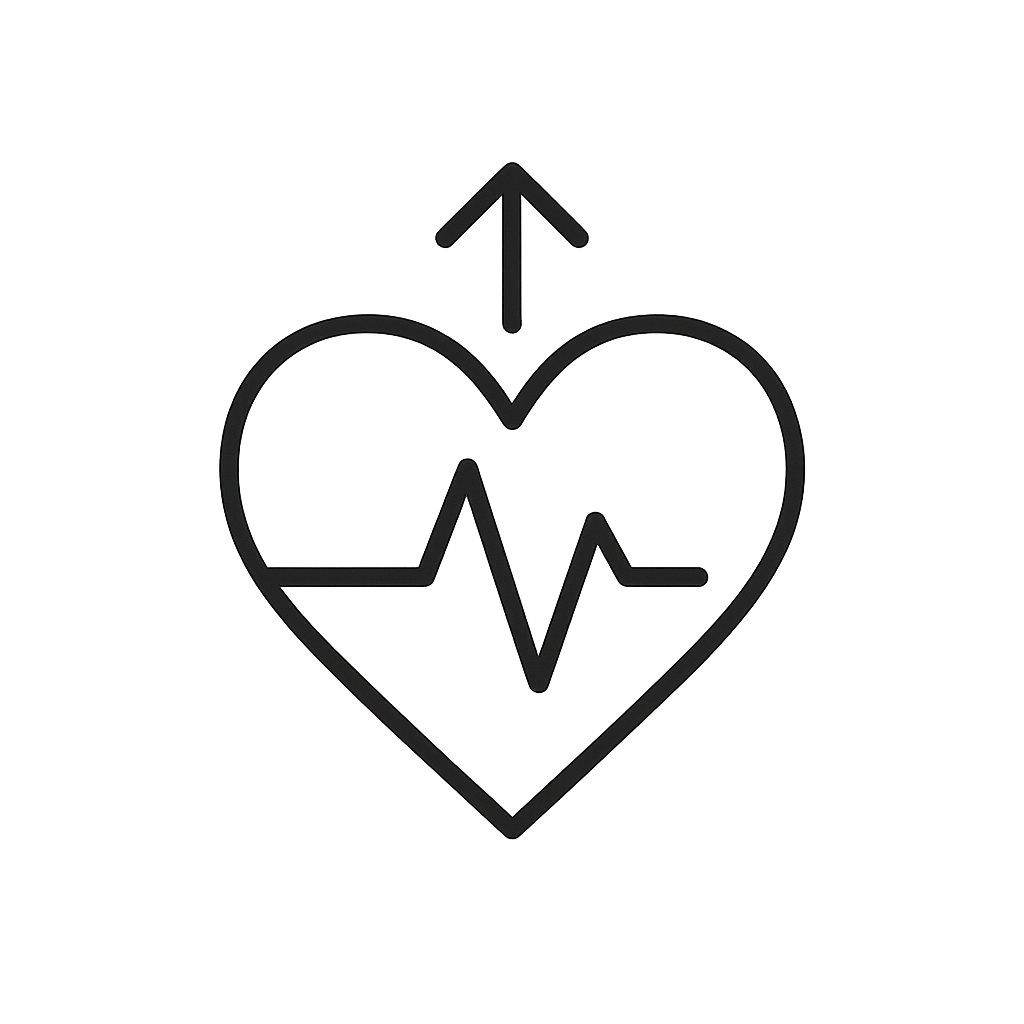}} \\
\midrule
GPT-5 \cite{openai_gpt5_systemcard_2025} & \textbf{65.42} & \textbf{53.44} & \textbf{63.66} & \textbf{61.11} & \textbf{64.61} & \textbf{57.39}\\
Gemini-2.5-Pro \cite{comanici2025gemini25} & \underline{54.47} & \underline{45.85} & \underline{55.53} & \underline{56.94} & \underline{59.07} & \underline{48.20} \\ \hline
Medgemma-4B \cite{sellergren2025medgemma} & 19.25 & 13.39 & 15.81 & 22.22 & 22.91 & 12.80\\
Lingshu-7B \cite{lasateam2025lingshu} & 32.21 & \text{22.93} & \text{31.74} & 36.11 & 37.45 & \underline{34.16}\\
Gemma3-4B \cite{team2025gemma} & 29.93 & 22.74 & 30.14 & \underline{38.89} & \text{37.58} & 21.74\\
Qwen2.5-VL-7B \cite{Qwen2.5-VL} & \underline{38.14} & \underline{27.06} & \underline{35.56} & \text{38.89} & \underline{41.06} & \text{31.43}\\
InternVL3.5-8B \cite{wang2025internvl3_5} & \text{32.39} & 21.23 & 26.24 & 31.94 & 33.98 & 24.97\\
Qwen3-VL-8B \cite{qwen3technicalreport} & \textbf{46.53} & \textbf{33.16} & \textbf{44.23} & \textbf{45.83} & \textbf{48.78} & \textbf{39.13} \\ \hline
Medgemma-27B \cite{sellergren2025medgemma} & 34.85 & 24.24 & 35.26 & 31.94 & 41.31 & 26.21\\
Lingshu-32B \cite{lasateam2025lingshu} & 38.59 & 26.92 & 33.37 & 40.28 & 41.83 & 32.05\\
Gemma3-27B \cite{team2025gemma} & \text{43.80} & \text{32.72} & \text{42.58} & \text{48.61} & \text{49.94} & 33.17\\     
Qwen2.5-VL-32B \cite{Qwen2.5-VL} & 42.34 & 28.84 & 37.92 & 47.22 & 46.33 & 33.29\\
Qwen2.5-VL-72B \cite{Qwen2.5-VL} & \textbf{52.28} & \underline{37.09} & \underline{48.07} & \underline{50.00} & \underline{51.87} & \textbf{43.23}\\
InternVL3.5-38B \cite{wang2025internvl3_5} & 43.25 & 31.76 & 37.51 & 33.33 & 44.79 & \text{38.88}\\
Qwen3-VL-30B-A3B \cite{qwen3technicalreport} & 42.70 & 32.31 & 39.68 & 41.67 & 46.33 & 33.54\\
Qwen3-VL-32B \cite{qwen3technicalreport} & \underline{50.91} & \textbf{40.33} & \textbf{49.54} & \textbf{52.78} & \textbf{53.93} & \underline{40.62} \\ \hline
InternVL3.5-241B-A28B \cite{wang2025internvl3_5} & \underline{56.11} & \underline{41.26} & \underline{50.93} & \underline{47.22} & \underline{57.40} & \underline{51.06}\\
Qwen3-VL-235B-A22B \cite{qwen3technicalreport} & \textbf{58.21} & \textbf{44.62} & \textbf{54.44} & \textbf{59.72} & \textbf{59.07} & \textbf{52.17} \\
\bottomrule
\end{tabular}

        }
}

\vspace{-3mm}     
\end{table} 
\section{Analysis}

\begin{figure*}[!htbp]
    \label{fig:ana}
    \centering

    \begin{subfigure}{0.28\textwidth}
        \includegraphics[width=\linewidth, height=0.22\textheight, keepaspectratio=false]{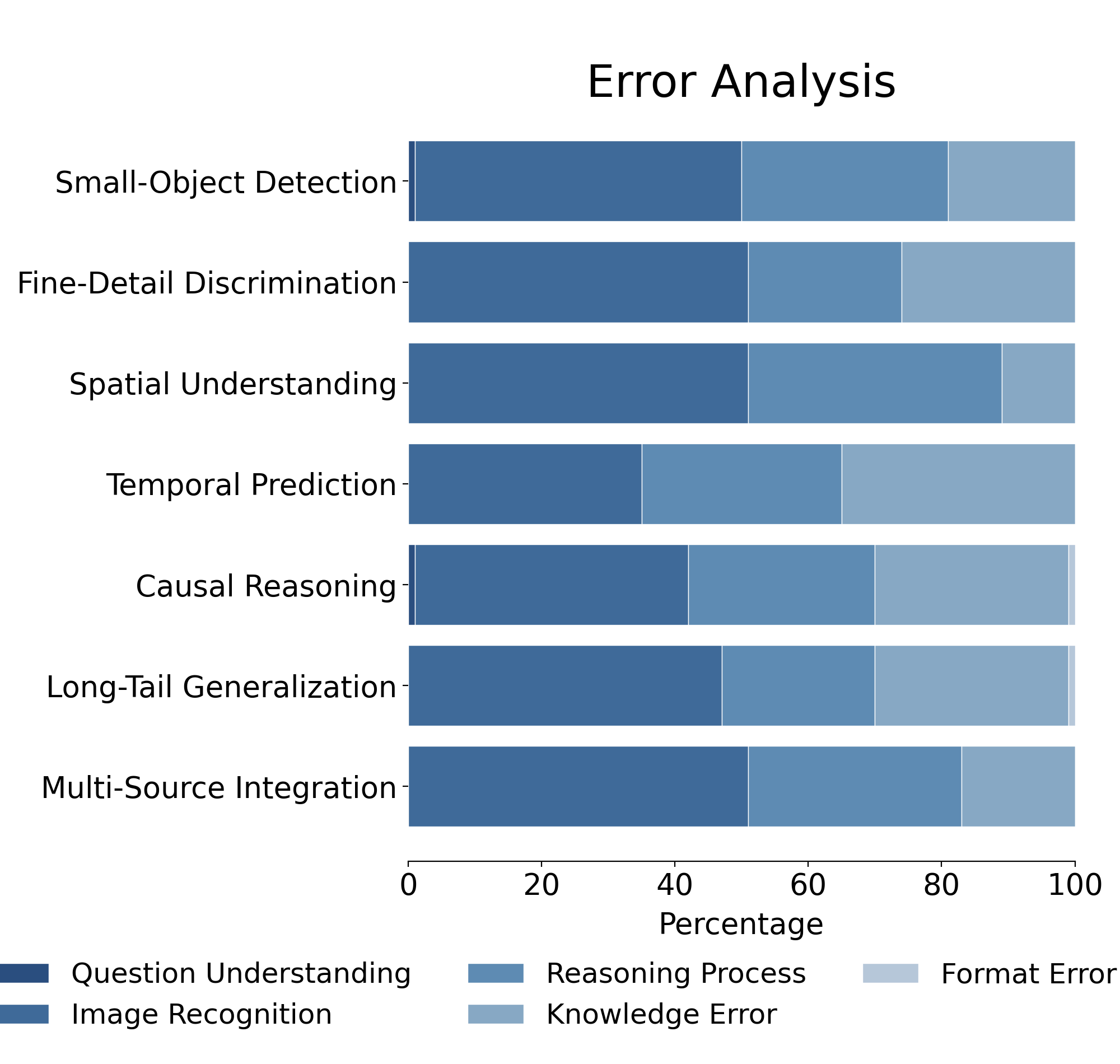}
        \caption*{(a)}
    \end{subfigure}
    \begin{subfigure}{0.23\textwidth}
        \includegraphics[width=\linewidth, height=0.22\textheight, keepaspectratio=false]{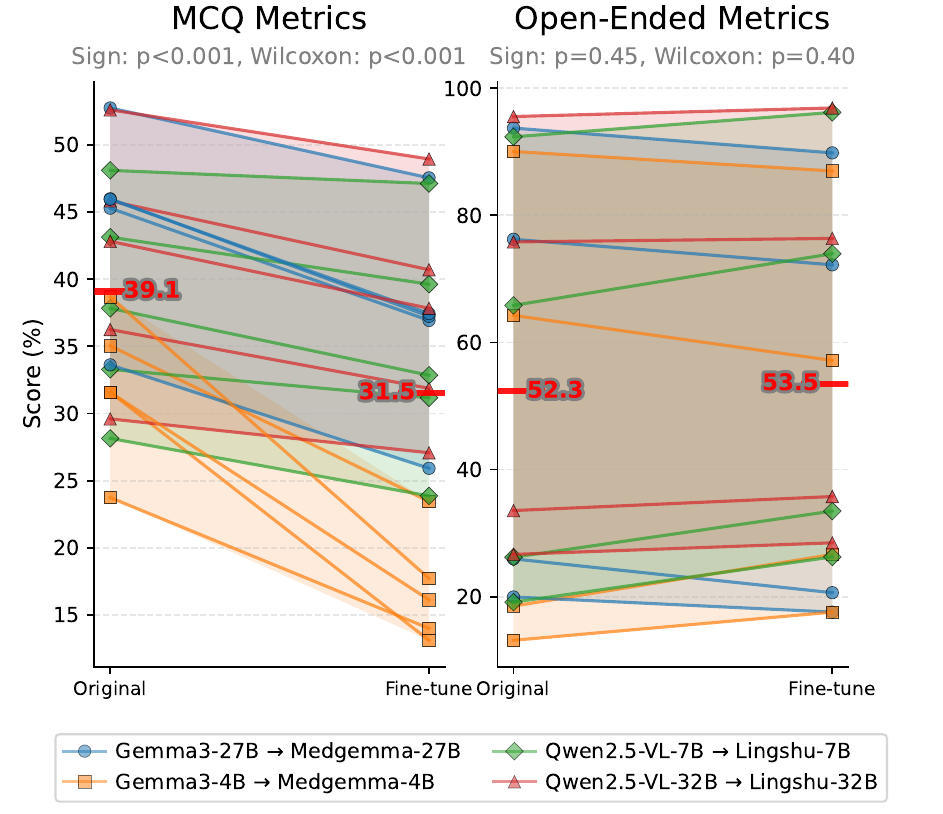}
        \caption*{(b)}
    \end{subfigure}
    \begin{subfigure}{0.23\textwidth}
        \includegraphics[width=\linewidth, height=0.22\textheight, keepaspectratio=false]{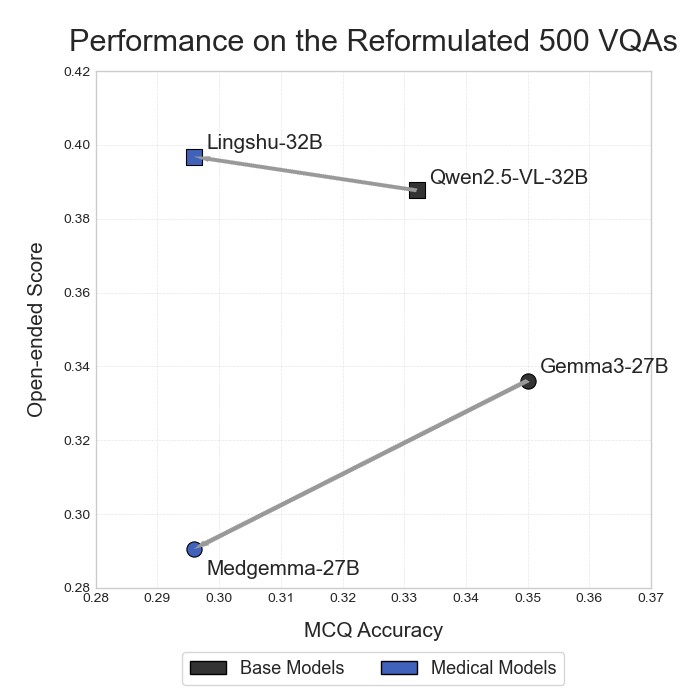}
        \caption*{(c)}
    \end{subfigure}
    \begin{subfigure}{0.24\textwidth}
        \includegraphics[width=\linewidth, height=0.22\textheight, keepaspectratio=false]{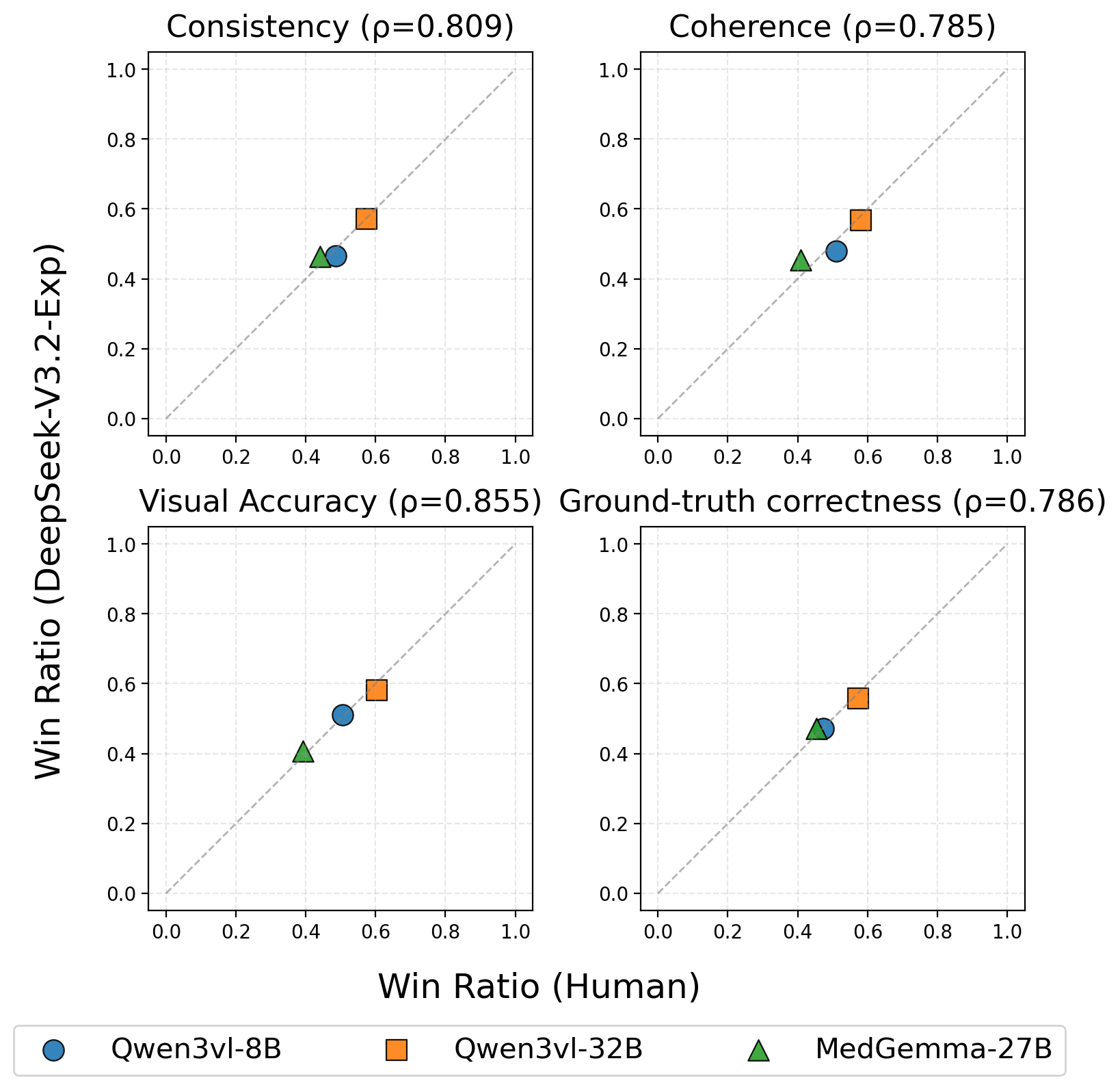}
        \caption*{(d)}
    \end{subfigure}
\vspace{-4mm}
    \caption{(a) Human-labeled GPT-5 error distribution across question dimensions. (b) Comparison of base models and corresponding medical models on the Med-CMR metrics. (c) MCQ and open-ended results from 500 reformulated MCQs for comparing base models and corresponding medical models. (d) Comparison of win ratios under human and LLM (DeepSeek-V3.2-Exp) evaluation across four dimensions. }
    \label{fig:benchmark_comparison}
\vspace{-3mm}
\end{figure*}

\subsection{How Far Are We from Expert-Level Complex Medical Reasoning?}
\vspace{-1mm}
To identify the remaining bottlenecks that prevent current MLLMs from achieving expert-level performance, we conduct a detailed error analysis of the best-performing model, GPT-5.
For each of the seven complexity dimensions, we sample 100 error cases and manually categorize them into five error types (recognition, reasoning, medical knowledge, question misunderstanding, and format issues). This allows us to isolate which capabilities most strongly limit current performance.

Overall, the dominant sources of error are image recognition, reasoning, and insufficient medical knowledge, while question understanding and format issues are rare. The relative proportions of these error types vary across dimensions, indicating that each type of challenge exposes distinct model limitations, as illustrated in Figure \hyperref[fig:ana]{4} (a). Detailed failure cases are provided in Supplementary Material.

\vspace{0.5em}

\noindent\textbf{Observation~1: Recognition-related errors form the largest portion of failure modes in visually demanding dimensions.}
GPT-5 often focuses on the overall appearance but neglects subtle details essential for reasoning, leading to missed small features, confusion of overlapping structures, and inconsistent tracking across slices. These issues occur most often in small-object detection, multi-source integration, fine-detail discrimination, and spatial understanding tasks, where subtle differences in shape, texture, or phase require precise localization. This suggests that current visual encoders lack robust multi-scale features and cross-frame consistency, leading the model to rely on coarse global cues rather than detailed local evidence.

\vspace{0.5em}

\noindent\textbf{Observation~2: Most reasoning failures arise from the model’s inability to connect evidence across views, time, and clinical context.} As a result, it often focuses on partial information and forms biased reasoning paths, leading to incorrect conclusions even when some observations are correctly identified. These errors are most pronounced in spatial understanding, multi-source integration, temporal prediction, and causal reasoning tasks, where incomplete evidence integration leads to biased conclusions, showing that the model’s reasoning often drifts toward partial information rather than maintaining overall coherence.

\vspace{0.5em}

\noindent\textbf{Observation~3: Even the most advanced model still struggles with questions requiring specialized medical knowledge.} These errors are most evident in temporal prediction, causal reasoning, and long-tail generalization tasks, where correct answers hinge on understanding disease mechanisms and clinically rare conditions that the model seldom encounters during training. This pattern indicates that further progress will require stronger medical knowledge integration and more robust generalization to rare or atypical cases.

\subsection{Are Medical MLLMs Better than General Ones?}
As shown in Figure \hyperref[fig:ana]{4 }(b), we observe a consistent pattern across all medically fine-tuned models: their performance on multiple-choice (MCQ) tasks declines compared with their respective base models, while on open-ended questions, the performance gap narrows and in some cases even reverses in favor of the medical models. In this section, we analyze the potential causes of this phenomenon.

We conduct a targeted error analysis to better understand the source of the performance gap. Specifically, we sampled 150 MCQ items that are correctly answered by Qwen2.5-VL-32B but missed by Lingshu-32B, and another 150 where Gemma-27B succeeded but Med-Gemma-27B failed. In addition, we examine 50 open-ended questions on which Lingshu-32B achieved higher scores than Qwen2.5-VL-32B. All samples are manually reviewed to characterize the underlying reasoning behaviors. 

For the MCQ tasks, we identify two major error patterns in medically fine-tuned models.
(1) The models often followed the correct diagnostic direction but overlooked subtle visual cues, resulting in incorrect answers.
(2) They frequently relied on pattern matching, directly linking a few salient features to prototypical diagnoses without integrating other contextual information. 
Together, these behaviors indicate a degradation of general multimodal reasoning ability after medical fine-tuning, where the models rely more on domain-specific associative patterns and less on cross-modal evidence integration. 

In contrast, for open-ended questions, the medical models often produced responses with richer medical semantics, although this pattern was not consistent across models. Even when fine-grained details were missed, they were more likely to generate text that matched the overall visual semantics or plausible diagnostic descriptions. One exception was Medgemma-27B, whose open-ended performance fell across all scoring dimensions, reflecting a marked degradation of its general reasoning ability.

To verify these observations, we reformulate 500 MCQs, originally favoring general models, into open-ended form and re-evaluated them using the same scoring criteria as the standard open-ended tasks. As shown in Figure \hyperref[fig:ana]{4} (c), Lingshu-32B achieved higher open-ended scores than its base model on these reformulated questions, whereas Medgemma-27B showed a decline relative to Gemma-27B.

In summary, medical fine-tuning allows models to acquire richer domain-specific semantics and generate more medically aligned responses, but it also causes a decline in their general multimodal reasoning ability. This decline is most evident in the handling of subtle visual and contextual details that are critical for precise discrimination and reliable reasoning. Consequently, medical models may underperform general models in tasks that demand fine-grained perception or complex clinical reasoning.

\subsection{Human Alignment Evaluation}
\label{sec:human}
In open-ended VQA evaluation, we adopt an LLM-as-a-Judge framework, where DeepSeek-V3.2-Exp serves as an external evaluator to provide standard-based scoring of model responses along four dimensions. To ensure the reliability of the LLM-based assessment, we conduct a systematic human-AI alignment analysis.

We randomly select 200 open-ended VQA samples and collect responses from three representative models: Qwen3-8B, Qwen3-32B~\cite{qwen3technicalreport}, and MedGemma-27B~\cite{sellergren2025medgemma}. Two annotators independently rank each model’s response on four dimensions — Consistency, Coherence, Visual Accuracy, and Ground-truth Correctness — following the same evaluation guidelines as the judging LLM, and then reconcile their rankings until reaching agreement. For each model pair, we calculate the win ratio, assigning scores of 1, 0.5, and 0 for win, tie, and loss, respectively, and aggregate the results to obtain the overall ratio. We further evaluate alignment by computing the Spearman correlation coefficient ($\bm{\rho}$) between human and LLM rankings to quantify alignment.

Figure \hyperref[fig:ana]{4} (d) illustrates the strong consistency between expert and LLM-based evaluations. Across all dimensions, the maximum difference in win ratio is only 0.0449, which occurs in the coherence dimension for MedGemma-27B. The correlation coefficients for consistency and visual accuracy exceed 0.8, while those for coherence and ground-truth correctness exceed 0.78. Together, these results demonstrate that our automated evaluation framework can serve as a reliable alternative to expert rating.
\vspace{-2mm}

\section{Conclusion}
We introduced \method, a fine-grained benchmark for evaluating complex medical reasoning in multimodal large language models. By decomposing reasoning into linked visual and inference tasks, covering diverse modalities and organs, and combining structured question design with multi-axis evaluation, Med-CMR provides a rigorous and clinically aligned assessment framework. Evaluating 18 state-of-the-art MLLMs reveals their strengths and persistent weaknesses, including remaining upper-bound gaps and the trade-offs introduced by medical fine-tuning. These findings offer concrete guidance for advancing medical MLLMs toward reliable clinical reasoning and for selecting models in real-world deployments.

\clearpage
\setcounter{page}{1}
\maketitlesupplementary

\section{Related Work}
\label{sec:related work}

\subsection{MLLMs in medical reasoning}
MLLMs show strong capability and great potential in medical clinical reasoning across visual and textual modalities. Existing MLLMs can be categorized into general-domain and medical-domain models, which differ in their data sources and specialization for medical reasoning tasks.

\noindent\textbf{General MLLMs.} 
Multimodal large language models have evolved from perception-driven systems to adaptive reasoning-centered frameworks \cite{yu2025vismem,yu2025visual,xing2025boosting,li2025unleashing}. Early general-domain MLLMs were mainly developed for multimodal perception by aligning visual encoders with pretrained language models.~\cite{li2023blip, liu2023visual, zhu2023minigpt}. Later models, including GPT-4~\cite{achiam2023gpt}, adopt end-to-end multimodal training and support richer modalities, leading to stronger general multimodal reasoning. Building on these architectures, reasoning-oriented models introduced explicit internal reasoning phases \cite{yu2025vismem, zhang2026chain, zhang2026pointcot, sun2026align}, where the model generates and refines intermediate reasoning traces before producing the final answer. Examples include OpenAI's o1 \cite{jaech2024openai} series, which allocates dedicated reasoning tokens. Unified frameworks such as Qwen3-VL \cite{qwen3technicalreport}, Gemini2.5 \cite{comanici2025gemini25}, and GPT-5 \cite{openai_gpt5_systemcard_2025} further integrate a dynamic “thinking” mode into general multimodal models, allowing adaptive control over the amount of internal reasoning based on task complexity.

\noindent\textbf{Medical MLLMs.}
Medical MLLMs are further trained on large-scale medical data and specifically developed for clinical and medical tasks. They have similarly evolved from a perception-aligned to a reasoning-enhanced paradigm. \cite{zhu2025pathology,zhu2026medeyes} Models such as Med-Flamingo \cite{moor2023med} and LLaVA-Med \cite{li2023llava} are among the first to extend general multimodal frameworks to the medical domain through medical data adaptation and multimodal alignment. More recent domain models, including LingShu \cite{lasateam2025lingshu} and Medgemma \cite{sellergren2025medgemma}, build upon medical semantic alignment and further incorporate multi-step reasoning data during training, enhancing implicit reasoning and factual consistency in medical applications.

\subsection{Multimodal Medical Benchmarks}
\label{related-benchmarks}
Multimodal medical benchmarks have evolved from early datasets focusing on simple perception and conceptual understanding to recent ones that begin to address complex clinical reasoning. Early medical VQA benchmarks, such as VQA-RAD \cite{lau2018dataset}, VQA-Med \cite{abacha2019vqa} , Path-VQA \cite{he2020pathvqa}, and SLAKE \cite{liu2021slake}, focus on evaluating recognition and factual understanding, aiming at perception-level comprehension rather than reasoning. Specifically, VQA-RAD focuses on radiology, VQA-Med covers multiple medical specialties, Path-VQA centers on pathology slides, and SLAKE provides bilingual annotations for broader accessibility. Later, large-scale and domain-diverse benchmarks were introduced to broaden modality coverage and improve generalization. PMC-VQA \cite{zhang2023pmc} collects image–text pairs from biomedical literature, OmniMedVQA \cite{hu2024omnimedvqa} covers multiple medical specialties, and GMAI-MMbench \cite{ye2024gmai} integrates heterogeneous data for general multimodal evaluation. These benchmarks enable large-scale assessment but still focus on shallow comprehension and retrieval-based reasoning. More recent reasoning-oriented benchmarks start to explore complex medical reasoning. MedXpertQA~\cite{zuo2025medxpertqa} contains a large number of questions that involve complex reasoning but was not specifically developed for this purpose. HIE-Reasoning \cite{baovisual} focuses on clinical complex reasoning using 133 neonatal MRI cases, which confines its scope to a narrow clinical setting. Both benchmarks indicate that current models have difficulty with complex reasoning, yet neither offers a fine-grained evaluation or systematic analysis of medical reasoning complexity. To bridge this gap, we present \method, the first benchmark that provides a fine-grained evaluation of MLLMs in complex clinical reasoning. To enable fine-grained evaluation, we decompose the complexity of clinical medical reasoning into three visual dimensions and four reasoning dimensions, each evaluated by a corresponding task.

\section{Case Studies}

\subsection{Case Studies Across Categories}
\vspace{6mm}

\begin{tcolorbox}[
    enhanced,
    title=Small-Object Detection
]

\begin{wrapfigure}{r}{0.38\linewidth}
    \vspace{-5mm}
    \includegraphics[width=\linewidth]{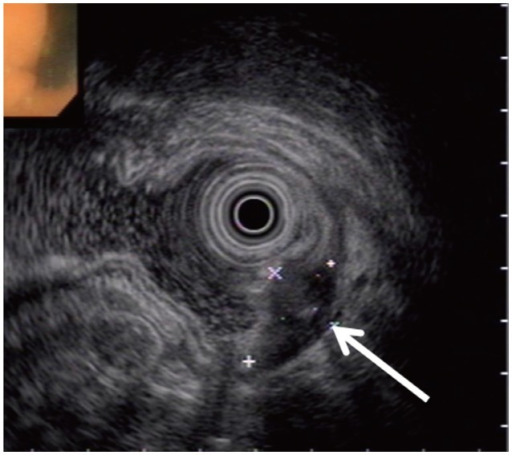}
\end{wrapfigure}

\textbf{Question: }Considering the mass's visual characteristics in the provided image and its relationship to surrounding structures, which feature best supports the conclusion that this is a pathologically significant tiny focus rather than an artifact or normal variation?

\textbf{Options: }A) The ovoid, heterogeneous hypoechoic focus with small echogenic foci centered in the submucosa and muscularis propria supports a true pathologic lesion.
B) The ovoid, heterogeneous hypoechoic focus with small central echogenic foci confined to the mucosal layer is diagnostic of a true pathologic lesion.
C) The ovoid, heterogeneous hypoechoic focus with central echogenic foci abutting only the muscularis mucosae suggests a focal pathology and therefore supports a true lesion rather than benign variation.
D) The ovoid, heterogeneous hypoechoic focus with small central echogenic foci confined to the mucosal layer and superficial submucosa is indicative of a true pathologic lesion rather than an artifact.
E) The small ovoid, heterogeneous hypoechoic focus with punctate echogenic foci localized predominantly to the mucosal and superficial submucosal layers is most consistent with a true pathologic lesion rather than an artifact.

\textbf{Answer: }A

\textbf{Explanation: }The mass shows an ovoid and heterogeneous hypoechoic focus with several small echogenic spots inside it. These features are centered in the submucosa and muscularis propria, which are deeper wall layers that do not normally contain such discrete structures. Artifacts on endoscopic ultrasonography usually do not form a stable ovoid shape, do not show internal echogenic foci arranged in a consistent pattern, and do not remain confined to specific wall layers. Normal variation of the gastric wall also does not create a sharply outlined hypoechoic mass in these deeper layers. Because the lesion has a defined shape, internal components, and a location that matches a true subepithelial abnormality, option A best supports the presence of a real small lesion rather than an artifact or normal tissue change.

\end{tcolorbox}

\vspace{12mm}

\begin{tcolorbox}[
    enhanced,
    title=Fine-Detail Discrimination
]

\begin{wrapfigure}{r}{0.28\linewidth}
    \vspace{-5mm}
    \includegraphics[width=\linewidth]{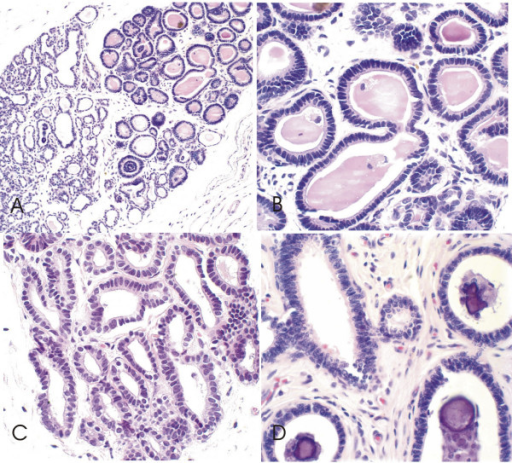}
\end{wrapfigure}

\textbf{Question: }Observe the overall pattern of the lesions in the tissue sections. Do the visual attributes suggest a central distribution or a pattern reflecting multicentric involvement?

\textbf{Answer: }The lesion pattern is consistent with multicentric involvement.

\textbf{Explanation: }The tissue sections show many separated glandular structures scattered throughout the sampled region. Each focus has nearly the same microscopic features, including a single layer of columnar epithelial cells, uniform nuclei, similar intraluminal secretions, and occasional calcifications. There is no visible gradient of change, no directional pattern, and no structure that can be identified as a central point of extension. Instead, the abnormalities appear as multiple small and discrete foci that share the same architecture. This pattern supports the interpretation that the lesions arise independently in multiple sites rather than spreading from one origin.
\end{tcolorbox}

\raggedbottom

\begin{tcolorbox}[
    enhanced,
    title=Spatial Understanding
]

\begin{wrapfigure}{r}{0.35\linewidth}
    \vspace{-5mm}
    \includegraphics[width=\linewidth]{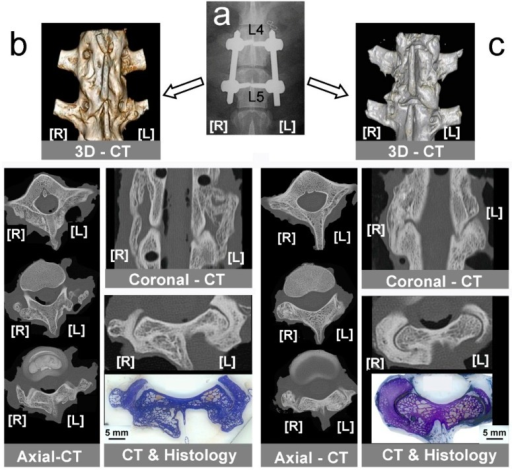}
\end{wrapfigure}

\textbf{Question: }Based on the comparison of structural and functional images in the affected region, how does the signal intensity correlate with the expected tissue viability in the reconstructed area?

\textbf{Options: }A) Signal intensity correlates with viable tissue on the left, reflecting optimal graft integration and successful fusion in the HA scaffold group.
B) Signal intensity aligns with viable tissue on the right, consistent with successful fusion.
C) Signal intensity corresponds to viable tissue on the left, indicating robust graft incorporation and successful fusion on the left side.
D) Signal intensity aligns with viable tissue on the left, consistent with successful fusion.
E) Signal intensity is greater on the left, consistent with preserved bone viability and fusion success localized to the left side.

\textbf{Answer: }B

\textbf{Explanation: }The CT images and histology sections show that only the right side has a continuous bridge of new bone. The right side appears denser on CT and matches the appearance of healthy, formed bone on the histology slice. The left side does not show the same continuous structure and has gaps instead of solid fusion. Because the structural views and the tissue views both point to good bone formation on the right side, the signal pattern matches viable tissue only on the right, which supports successful fusion on that side.
\end{tcolorbox}

\vspace{12mm}

\begin{tcolorbox}[
    enhanced,
    title=Temporal Prediction
]

\begin{wrapfigure}{r}{0.36\linewidth}
    \vspace{-5mm}
    \includegraphics[width=\linewidth]{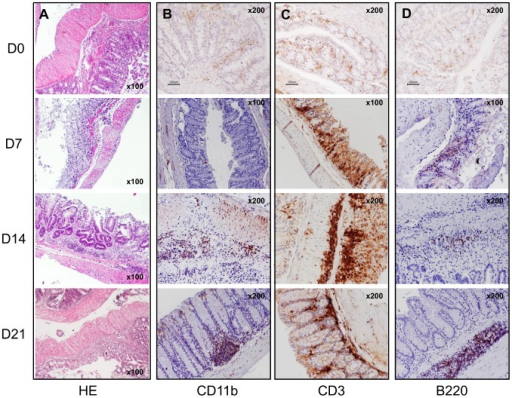}
\end{wrapfigure}

\textbf{Question: }What visual configuration most strongly indicates a therapy-related process rather than a baseline condition in the affected region across the different time-points?

\textbf{Answer: }Progressive accumulation of CD3+, followed by B220+ and Mac-1+ inflammatory cells at later time-points.

\textbf{Explanation: }The images show a time-dependent change in the inflammatory pattern of the colon. At day 0, the tissue architecture is mostly preserved and immune cell staining is sparse. By day 7, CD3\textsuperscript{+} T cells begin to accumulate along the mucosa, indicating an early lymphocytic response. At day 14, CD3\textsuperscript{+} staining becomes dense and widespread, showing that T-cell infiltration is the dominant feature at this stage. At the same time, B220\textsuperscript{+} B cells start to appear, although at a lower level. By day 21, the inflammatory infiltrate becomes more mixed. B220\textsuperscript{+} B cells and Mac-1\textsuperscript{+} myeloid cells increase, filling deeper layers of the mucosa and submucosa. This progression in the order of appearance---first CD3\textsuperscript{+} T cells, then B220\textsuperscript{+} B cells, and finally Mac-1\textsuperscript{+} cells---shows a dynamic sequence instead of a baseline condition. This temporal pattern is consistent with a therapy-related or injury-related immune process, where different immune cell populations are recruited at different stages.

\end{tcolorbox}

\begin{tcolorbox}[
    enhanced,
    title=Causal Reasoning
]

\begin{wrapfigure}{r}{0.35\linewidth}
    \vspace{-5mm}
    \includegraphics[width=\linewidth]{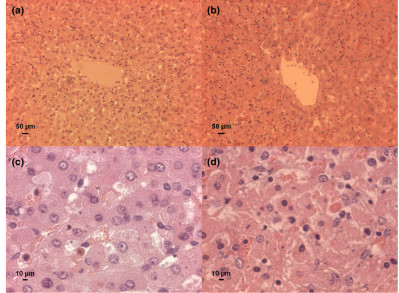}
\end{wrapfigure}

\textbf{Question: }What visual configuration most strongly indicates a therapy-related rather than baseline process in the affected region?

\textbf{Answer: }Sinusoidal infiltration by polymorphonuclear neutrophils and lymphocytes in the centrolobular liver

\textbf{Explanation: }The images compare two conditions that differ only by the use of the recruitment manoeuvre: pressure-controlled ventilation alone (PCV) and pressure-controlled ventilation plus recruitment (PCV+R). In the PCV group (a, c), centrolobular liver tissue shows largely preserved architecture with only a few scattered inflammatory cells in the sinusoids. In the PCV+R group (b, d), the same region of the liver now shows dense sinusoidal infiltration by polymorphonuclear neutrophils and lymphocytes. Because the animals, organ, stain, and magnification are otherwise matched, this new inflammatory pattern appears specifically in the group that received the additional manoeuvre. The most reasonable causal interpretation is that the recruitment manoeuvre is associated with therapy-related liver injury, and the visual configuration that captures this causal effect is the sinusoidal infiltration by neutrophils and lymphocytes in the centrolobular area.
\end{tcolorbox}

\vspace{12mm}

\begin{tcolorbox}[
    enhanced,
    title=Long-Tail Generalization
]

\begin{wrapfigure}{r}{0.35\linewidth}
    \vspace{-5mm}
    \includegraphics[width=\linewidth]{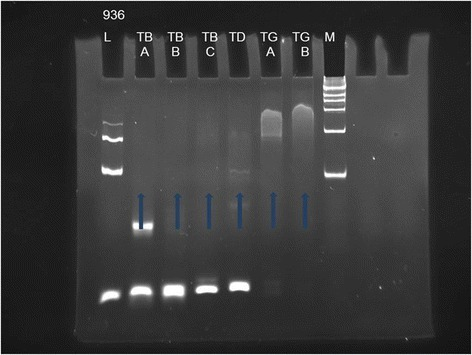}
\end{wrapfigure}

\textbf{Question: }Which feature in the observed pattern most supports an unusual mechanism rather than a common diagnosis?

\textbf{Options: }A) Germline configuration of T-cell receptor loci interpreted as evidence for a de-differentiated T-cell neoplasm rather than an NK-cell origin.
B) Germline configuration of T-cell receptor genes taken to exclude an NK-cell origin and instead indicate a cryptic T-cell clone with non-productive rearrangements.
C) Germline T-cell receptor configuration definitively confirming lineage infidelity as a de-differentiated T-cell neoplasm rather than a potential NK-cell origin.
D) Germline T-cell receptor loci interpreted as definitive evidence of lineage infidelity consistent with a de-differentiated T-cell neoplasm rather than an NK-cell lineage.
E) Germline T-cell receptor configuration indicating a potential NK-cell origin.

\textbf{Answer: }E

\textbf{Explanation: }The gel shows that all tested samples have a germline pattern for the T-cell receptor (TCR) loci rather than a discrete clonal rearranged band. In other words, the TCR genes remain in their unrearranged configuration, similar to the control ladder, and there is no evidence of a clonal T-cell population. True T-cell lymphomas almost always show rearranged TCR genes, so a purely germline configuration argues against a conventional T-cell neoplasm. Natural killer (NK) cells do not rearrange TCR genes, so finding germline TCR in a lymphoid malignancy strongly supports an NK-cell origin. This unusual lineage assignment, based on the germline TCR configuration, is the key feature that points to a rare NK-cell lymphoma mechanism rather than a common T-cell lymphoma diagnosis.
\end{tcolorbox}

\begin{tcolorbox}[
    enhanced,
    title=Multi-source integration
]

\begin{wrapfigure}{r}{0.24\linewidth}
    \vspace{-5mm}
    \includegraphics[width=\linewidth]{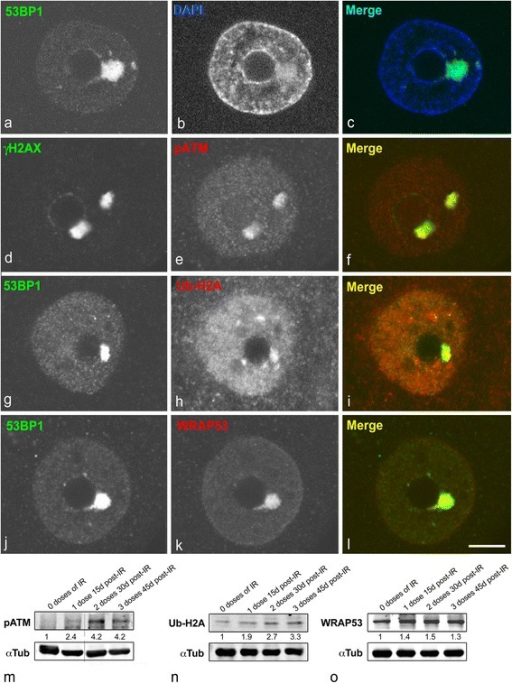}
\end{wrapfigure}

\textbf{Question: }Which visual feature in the image series most strongly supports the conclusion that the observed lesions reflect a coordinated pathological process rather than random occurrences?

\textbf{Answer: }The recurring colocalization of multiple DNA damage and repair markers within the same PDDF in the affected nucleus.

\textbf{Explanation: }The images show that several DNA damage--related markers --- 53BP1, $\gamma$H2AX, pATM, Ub-H2A, and WRAP53 --- are all concentrated within the same intranuclear structure, the persistent DNA damage focus (PDDF). In every panel, these markers appear in the same region of the nucleus, demonstrating repeated colocalization rather than scattered or independent lesions. This stable and shared localization pattern indicates that the abnormalities arise from a single coordinated DNA damage process instead of multiple random events. The Western blot data support this by showing IR-dependent increases in these same repair markers, matching the focal nuclear accumulation observed microscopically.

\end{tcolorbox}

\begin{center}
\large \textbf{Med-CMR examples across seven categories.}
\end{center}

\vspace{6mm}

\subsection{GPT-5 Error Case Studies}
\vspace{6mm}

\begin{tcolorbox}[
    enhanced,
    title=Question Understanding
]

\begin{wrapfigure}{r}{0.35\linewidth}
    \vspace{-5mm} 
    \includegraphics[width=\linewidth]{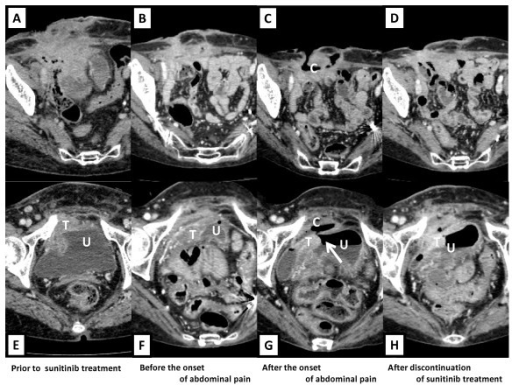}
\end{wrapfigure}

\textbf{Question:} Considering the affected region's relationship with adjacent structures, which aspect most clearly supports the significance of the subtle finding?

\textbf{Options:}  
A) A focal mural discontinuity at the posterior bladder wall contiguous with the tumor, consistent with a tiny bladder dome perforation.
B) A minute defect at the inferior (neck) aspect of the anterior bladder wall adjacent to the lesion suggesting intraperitoneal communication.
C) A small defect at the anterior bladder wall adjacent to the lesion.
D) A minute defect at the inferior (neck) aspect of the anterior bladder wall adjacent to the lesion suggesting intraperitoneal communication.
E) A minute defect at the lateral anterior bladder dome adjacent to the lesion.

\textbf{Correct Answer: }{\color{black} C}

\textbf{Model Answer: }{\color{orange!90!black} E}



\textbf{Model Reasoning: }{\color{blue!40!black}The question asks which imaging detail, based on how the lesion relates to nearby structures, best confirms the clinical importance of a subtle CT finding. Serial axial CT images show a pelvic mass abutting the urinary bladder. At the lateral anterior bladder dome, right where the mass contacts the bladder wall, a minute defect in the bladder wall at that site. This focal defect at the dome adjacent to the lesion most strongly indicates a small but clinically important breach or early communication. Other observations either occur at locations that do not relate to the dome or do not show this discrete dome defect, so they lack the same direct anatomic relevance.}

\textbf{Error Interpretation:} The model did not ground its interpretation in the specific structural detail that the question was designed to assess. It shifted the abnormality toward a dome-based location and responded as if the finding were meant to highlight a lateral dome defect, which is not what the question described. By doing so, it effectively answered a different question than the one asked. This indicates a question understanding error.

\end{tcolorbox}

\begin{tcolorbox}[
    enhanced,
    title=Image Recognition
]

\begin{wrapfigure}{r}{0.4\linewidth}
    \vspace{-5mm} 
    \includegraphics[width=\linewidth]{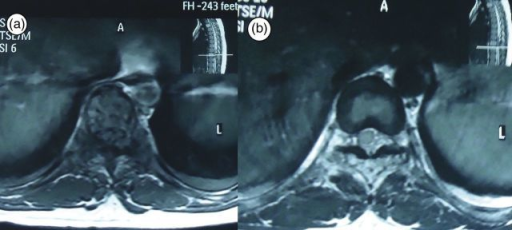}
\end{wrapfigure}

\textbf{Question: }Analyze the relationship between the lesions and their surrounding anatomical structures. Are the lesions primarily acting as a primary process with satellites or are they indicative of secondary spread from another region?

\textbf{Options: }A) The pattern suggests a primary paraspinal soft-tissue neoplasm centered in the paraspinal musculature with contiguous spread into adjacent vertebrae and focal epidural impingement rather than a primary vertebral lesion with satellite deposits.
B) The pattern suggests multifocal paravertebral soft tissue disease with superficial secondary vertebral scalloping and partial epidural compression, indicating primary soft-tissue origin with subsequent superficial bony remodeling rather than true bony invasion.
C) The pattern suggests a primary vertebral lesion with satellite involvement, demonstrating extraosseous extension and epidural encasement of the cord.
D) The pattern suggests a primary vertebral process with multifocal satellite lesions limited to the paravertebral soft tissues without true epidural encasement of the thecal sac.
E) The pattern suggests multifocal paravertebral soft tissue disease with secondary involvement of the vertebral body margins and partial epidural displacement of the thecal sac without true vertebral origin.

\textbf{Correct Answer: }{\color{black} C}

\textbf{Model Answer: }{\color{orange!90!black} B}



\textbf{Model Reasoning: }{\color{blue!40!black}The task is to decide whether the observed spinal lesions originate primarily from paravertebral soft tissues with subsequent spread to adjacent structures such as the bony vertebra or epidural space, or whether they represent secondary involvement of soft tissues due to a primary vertebral process. Axial MR images at the thoracic level show bulky, lobulated soft tissue masses in the paravertebral regions. The masses abut and indent the posterolateral vertebral body and pedicle with cortical indentation and shallow scalloping rather than a marrow centered expansile lesion. There is extension toward the epidural space causing partial effacement of the thecal sac and cord compression. Intervertebral disc contours appear relatively preserved, and involvement is asymmetric and multifocal along the paravertebral gutters. Primary vertebral pathology typically begins within the vertebral body marrow with centripetal expansion, endplate and disc involvement, and then secondarily extends into paravertebral or epidural spaces, but those features such as marrow centered destruction and disc or endplate collapse are not dominant here. The conspicuous finding is large paravertebral soft tissue masses that contact the bone surface and produce scalloping-type superficial remodeling, implying outside-in influence rather than true primary osseous destruction. Multifocal paravertebral lesions with only superficial secondary cortical scalloping and epidural encroachment favor a soft tissue origin with satellites rather than a primary osseous process. Preservation of disc spaces further argues against primary spondylodiscitis-like or marrow-origin lesions. The pattern suggests multifocal paravertebral soft tissue disease with superficial secondary vertebral scalloping and partial epidural compression, indicating primary soft-tissue origin with subsequent superficial bony remodeling rather than true bony invasion.}

\textbf{Error Interpretation: }The model did not correctly identify the true site of origin of the lesions. The expected pattern reflects a primary vertebral process with secondary extraosseous extension, but the model interpreted the appearance as soft-tissue–dominant disease with only superficial effects on the vertebrae. This shows that the model misread the structural relationships on the images. The error comes from image recognition.

\end{tcolorbox}


\begin{tcolorbox}[
    enhanced,
    title=Reasoning Process
]

\begin{wrapfigure}{r}{0.35\linewidth}
    \vspace{-5mm} 
    \includegraphics[width=\linewidth]{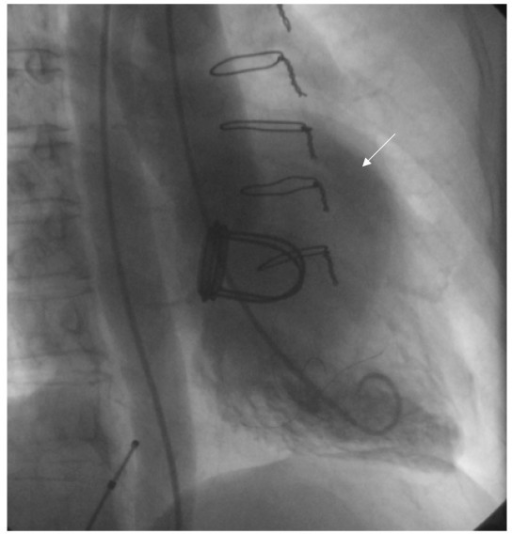}
\end{wrapfigure}

\textbf{Question: }What visual configuration in the image most strongly indicates a therapy-related adverse effect rather than a baseline process?

\textbf{Options: }A) Displacement and extrinsic narrowing of the left circumflex artery by an adjacent aneurysm.
B) External indentation of the left circumflex artery by an adjacent aneurysmal sac accompanied by focal vessel irregularity and intraluminal filling defect consistent with intrinsic luminal disease rather than pure extrinsic compression.
C) Displacement of the left circumflex artery by an adjacent aneurysm with preserved luminal caliber and absence of any vessel displacement or extrinsic narrowing, indicating post-procedural anatomical remodeling rather than acute compression.
D) External indentation of the left circumflex artery by an adjacent aneurysmal sac without appreciable vessel indentation or displacement, suggesting no compression.
E) Displacement of the left circumflex artery by an adjacent aneurysm with preserved luminal caliber and no adjacent aneurysmal sac, suggesting intrinsic vessel tortuosity rather than extrinsic narrowing.

\textbf{Correct Answer: }{\color{black} A}

\textbf{Model Answer: }{\color{orange!90!black} B}



\textbf{Model Reasoning: }{\color{blue!40!black}The task is to identify which visual pattern on the image points to a complication caused by therapy (iatrogenic or extrinsic effect) rather than a patient’s baseline intrinsic coronary disease. The fluoroscopic cardiac image shows surgical hardware such as a ring or annuloplasty device and wires, indicating prior cardiac intervention. The arrow points to a coronary segment along the lateral wall. At the arrowed site the coronary contour shows a sharp outward concavity adjacent to a rounded external opacity consistent with an aneurysmal or pseudoaneurysmal sac. In addition to this external impression, the vessel segment at the point of contact appears irregular with a focal intraluminal filling defect, implying that the wall and lumen themselves have been altered rather than simply pushed aside. Baseline intrinsic coronary disease such as stable atherosclerosis usually appears as smooth luminal narrowing without an associated adjacent sac or combined pattern of external indentation plus intraluminal defect. A therapy-related adverse effect, such as a postoperative pseudoaneurysm with associated intraluminal thrombus, can both indent the artery from outside and create focal luminal irregularity. This combined appearance of an external aneurysmal sac, focal indentation, and intraluminal filling defect is best captured by option B: external indentation of the left circumflex artery by an adjacent aneurysmal sac accompanied by focal vessel irregularity and intraluminal filling defect consistent with intrinsic luminal disease rather than pure extrinsic compression.}

\textbf{Error Interpretation: }The model correctly identifies the adjacent aneurysmal structure and its relationship to the left circumflex artery, but it misinterprets the vessel contour and luminal appearance during its reasoning process. While working through the options, it introduces features such as focal vessel irregularity and an intraluminal filling defect, even though these are not supported by the image. This overinterpretation leads it away from the actual pattern of displacement and extrinsic narrowing that indicates a therapy-related effect. The mistake therefore stems from a flawed reasoning process in which unwarranted details are added and the visual findings are overstated, causing the model to favor an option that does not match the observed configuration.
\end{tcolorbox}


\begin{tcolorbox}[
    enhanced,
    title=Domain Understanding Error
]

\begin{wrapfigure}{r}{0.4\linewidth}
    \vspace{-5mm}
    \includegraphics[width=\linewidth]{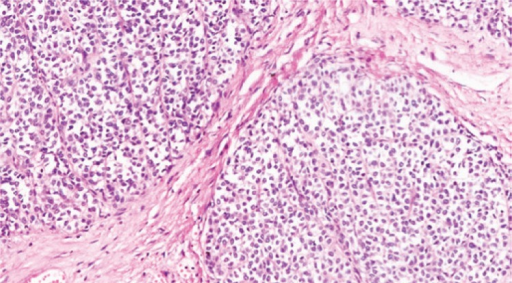}
\end{wrapfigure}

\textbf{Question: }Analyze the microscopic pattern within the affected region. What unusual feature could suggest a rare disease process rather than a common condition?

\textbf{Options: }A) Epithelioid cells with mildly irregular nuclei and low mitotic activity in an organoid pattern, with isolated tiny foci of degenerative change mischaracterized as necrosis.
B) Irregular nuclei with infrequent mitoses in a disorganized architectural pattern, with diffuse inflammation without necrotic foci.
C) Epithelioid cells with irregular nuclei and low mitotic activity in an organoid/epithelioid architectural pattern, with small confluent areas interpreted as ischemic-type necrosis absent.
D) Irregular nuclei with infrequent mitoses in an epithelioid/organoid arrangement, with scant focal apoptosis without established coagulative necrosis.
E) Irregular nuclei with numerous mitoses in an organoid/epithelioid architectural pattern, including focal necrosis.

\textbf{Correct Answer: }{\color{black} E}

\textbf{Model Answer: }{\color{orange!90!black} A}



\textbf{Model Reasoning: }{\color{blue!40!black}The task is to examine the histologic image and identify an uncommon microscopic feature that would point toward a rare disease process rather than a typical condition. The field shows solid nests or sheets of uniform epithelioid cells separated by fibrous septa. The nuclei are mildly irregular, the chromatin is fairly even, and there are only rare mitotic figures indicating low proliferative activity. The architecture is organoid or nested. There are tiny pale foci suggestive of degenerative change but not broad, established necrosis. Common reactive or inflammatory conditions would show mixed inflammatory cells and lack an organoid nesting pattern with even this low level of mitotic activity. Many common malignant processes would show marked pleomorphism, high mitotic rates, and overt coagulative necrosis, none of which is prominent here. The combination of epithelioid cytology, organoid growth, and low mitotic activity with only minute degenerative foci is more aligned with a rare epithelioid or neuroendocrine-type process rather than a common pathology. This pattern is best captured by option A: epithelioid cells with mildly irregular nuclei and low mitotic activity in an organoid pattern, with isolated tiny foci of degenerative change mischaracterized as necrosis.}

\textbf{Error Interpretation: }The model’s mistake stems from a gap in its pathological knowledge. Although the image shows epithelioid cells in an organoid pattern with mildly irregular nuclei, low mitotic activity, and only tiny foci of degenerative change, the model did not recognize that this specific combination is more suggestive of certain rare epithelioid or neuroendocrine-type processes rather than of a common condition. Instead, it interpreted these findings as nonspecific changes without appreciating their diagnostic implication. This indicates that the model lacks the domain knowledge required to distinguish the histologic pattern of a rare disease from patterns seen in more typical lesions.

\end{tcolorbox}

\begin{tcolorbox}[
    enhanced,
    title=Format Error
]

\begin{wrapfigure}{r}{0.35\linewidth}
    \vspace{-5mm} 
    \includegraphics[width=\linewidth]{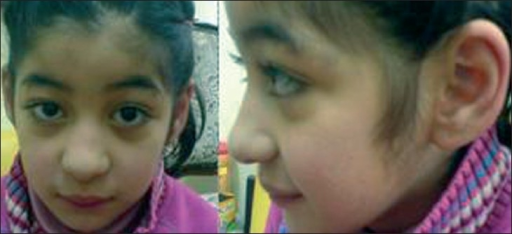}
\end{wrapfigure}

\textbf{Question: }By examining the visible face features, can you deduce the nature of the developmental anomaly and how it affects facial structure organization?

\textbf{Options: }A) The facial features indicate a developmental anomaly characterized by frontal hair upsweep, long face, high-set ears, sparse eyebrows, long eyelashes, epicanthic folds, and sparse eyebrows, reflecting altered craniofacial organization.
B) The facial features suggest a developmental anomaly with frontal hair upsweep, short face, high-set ears, sparse eyebrows, short eyelashes, and absent epicanthic folds, indicating altered craniofacial organization.
C) The facial features suggest a developmental anomaly with frontal hair upsweep, long face, high-set ears, sparse eyebrows, short eyelashes, epicanthic folds, and sparse eyebrows, indicating altered craniofacial organization.
D) The facial features indicate a developmental anomaly characterized by frontal hair upsweep, short face, low-set ears, thick eyebrows, long eyelashes, and absent epicanthic folds, reflecting altered craniofacial organization.
E) The facial features indicate a developmental anomaly characterized by frontal hair upsweep, long face, low-set ears, thick eyebrows, long eyelashes, epicanthic folds, and full eyebrows, reflecting altered craniofacial organization.

\textbf{Correct Answer: }{\color{black} E}

\textbf{Model Answer: }{\color{orange!90!black} /S}



\textbf{Model Reasoning: }{\color{blue!40!black}The question asks to infer, from the facial features shown, what developmental anomaly is present and how it alters the organization of facial structures. The child’s face shows a frontal hair upsweep with an elongated facial contour, high-set external ears, sparse eyebrows, long eyelashes, and the presence of epicanthic folds. These findings occur bilaterally and affect multiple craniofacial units. The combination of facial elongation, altered ear position, reduced eyebrow density, eyelash prominence, and epicanthic folds points to a disturbance in craniofacial development that affects the proportional relationships and patterned organization of the face. Such abnormalities reflect altered growth or alignment of craniofacial elements derived from early embryologic processes governing facial morphology. Mechanistically, disruption in the coordinated development of these structures leads to an atypical arrangement of facial features, producing the observed craniofacial organization.}

\textbf{Error Interpretation: }The model output “/S” instead of a valid option due to a breakdown in its answer formatting logic, likely caused by confusion between text generation and option selection steps during decoding.

\end{tcolorbox}

\begin{center}
\large \textbf{GPT-5 error examples across five error types.}
\end{center}

\vspace{6mm}

\section{Data Leakage Analysis}
To ensure that Med-CMR provides a fair and reliable evaluation of multimodal models, we analyze the dataset for potential leakage signals. This analysis shows that Med-CMR exhibits minimal evidence of leakage.

During benchmark curation, we follow a controlled process designed to reduce the chance of overlap with existing corpora. All questions are generated by instantiating human-designed templates that are created manually and adjusted only slightly to match each case, rather than adapted from any external materials. Correct answers are derived from the caption content and rewritten in a new form. For multiple-choice items, distractors are proposed by several models and then refined through human selection. This pipeline greatly reduces the chance of data leakage.

We further examine potential contamination using N-gram Accuracy, specifically ROUGE-L \cite{lin2004rouge} and edit distance similarity, which have been adopted in earlier studies \cite{xu2024benchmarking}. We sample 2,000 MCQs and 500 open-ended questions and convert the MCQs into open-ended form to collect free-text predictions so that lexical similarity can be measured directly. We evaluate three top-performing models, including both proprietary and open-source models, since these models offer a representative upper bound for potential overlap. We compare each model’s responses with the reference answers to obtain the ROUGE-L and edit distance similarity, and the results are reported in Table \hyperref[tab:1]{1} and Table \hyperref[tab:2]{2}. Using the criteria introduced in prior work \cite{xu2024benchmarking}, the measured similarity is far below the level associated with contamination, indicating that the leakage risk is minimal.

\begin{table}[t]
\label{tab:1}
\centering
\begin{minipage}{0.55\linewidth}
\centering
\begin{tabular}{lccc}
\hline
\textbf{Model} 
& \textbf{MCQ-converted} 
& \textbf{OE} 
& \textbf{Overall} \\
\hline
GPT-5 \cite{openai_gpt5_systemcard_2025}
& 0.123 & 0.132 & 0.124 \\

InternVL3.5-241B-A28B \cite{wang2025internvl3_5}
& 0.073 & 0.108 & 0.079 \\

Qwen3-VL-235B-A22B \cite{qwen3technicalreport}
& 0.047 & 0.072 & 0.051 \\
\hline
\end{tabular}
\caption{ROUGE-L (Longest Common Subsequence F-score) across MCQ-converted, open-ended, and overall subsets.}
\end{minipage}
\end{table}

\begin{table}[t]
\label{tab:2}
\centering
\begin{minipage}{0.55\linewidth}
\centering
\begin{tabular}{lccc}
\hline
\textbf{Model} 
& \textbf{MCQ-converted} 
& \textbf{OE} 
& \textbf{Overall} \\
\hline
GPT-5 \cite{openai_gpt5_systemcard_2025}
& 0.069 & 0.078 & 0.071 \\

InternVL3.5-241B-A28B \cite{wang2025internvl3_5}
& 0.049 & 0.068 & 0.052 \\

Qwen3-VL-235B-A22B \cite{qwen3technicalreport}
& 0.032 & 0.050 & 0.035 \\
\hline
\end{tabular}
\caption{Edit distance similarity across MCQ-converted, open-ended, and overall subsets.}
\end{minipage}
\end{table}

\section{Medical Coverage and Distribution}

\subsection{Modality}

\begin{table}[h!]
\centering
\begin{tabularx}{0.5\linewidth}{X m{1.4cm} m{0.85cm}}
\hline
\textbf{Modality} & \textbf{Count} & \textbf{Prop.} \\
\hline
Pathology / Microscopy & 5662 & 0.264 \\
Photography / Dermatology & 4662 & 0.218 \\
CT & 3441 & 0.161 \\
X-ray \& Fluoroscopy & 2593 & 0.121 \\
MRI & 2017 & 0.094 \\
Ultrasound & 896 & 0.042 \\
Endoscopy & 764 & 0.036 \\
Ophthalmology / Optical & 649 & 0.030 \\
Nuclear Medicine / PET & 436 & 0.020 \\
Physiologic Signals & 225 & 0.011 \\
Radiotherapy & 54 & 0.003 \\
Derived / Meta & 23 & 0.001 \\
\hline
\end{tabularx}
\caption{Modality coverage list and distribution. Prop. denotes proportion.}
\end{table}

\subsection{Body System}

\begin{table}[h!]
\centering
\begin{tabularx}{0.5\linewidth}{X m{1.5cm} m{1.0cm}}
\hline
\textbf{Body System} & \textbf{Count} & \textbf{Prop.} \\
\hline
Digestive & 3843 & 0.186 \\
Integumentary & 3218 & 0.156 \\
Nervous & 2545 & 0.123 \\
Skeletal & 2266 & 0.110 \\
Other / NA & 1848 & 0.089 \\
Respiratory & 1824 & 0.088 \\
Cardiovascular & 1821 & 0.088 \\
Reproductive & 1274 & 0.062 \\
Muscular & 842 & 0.041 \\
Urinary & 813 & 0.039 \\
Endocrine & 360 & 0.017 \\
\hline
\end{tabularx}
\caption{Body system coverage list and distribution. Prop. denotes proportion.}
\end{table}

\section{Task Mapping}

\renewcommand{\arraystretch}{1.2}
\begin{table}[h!]
\centering
\begin{tabularx}{0.8\linewidth}{X X m{1.2cm}}
\hline
\textbf{Complexity Dimension} & \textbf{Corresponding Task} & \textbf{Abbr.} \\
\hline
\multicolumn{3}{l}{\textbf{Visual Complexity}} \\
\hline
Small-Object Detection & Small-Object Detection & SOD \\
Fine-Detail Discrimination & Fine-Detail Discrimination & FDD \\
Spatial Understanding & Spatial Understanding & SU \\
\hline
\multicolumn{3}{l}{\textbf{Reasoning Complexity}} \\
\hline
Temporal Prediction & Temporal Prediction & TP \\
Causal Reasoning & Causal Reasoning & CR \\
Long-Tail Generalization & Long-Tail Generalization & LTG \\
Multi-Source Integration & Multi-Source Integration & MSI \\
\hline
\end{tabularx}
\caption{Mapping between multimodal medical reasoning complexity dimensions and tasks. Abbr. denotes task abbreviation.}
\end{table}

\section{Prompts}

\subsection{MCQ Response Generation Prompt}

\begin{table}[h]
\centering
\setlength{\abovecaptionskip}{3pt}

\begin{tcolorbox}[
    colback=gray!15,
    colframe=black,
    arc=1.3mm,
    boxrule=0.8pt,
    width=0.95\linewidth,
    left=10pt,
    right=10pt,
    top=10pt,
    bottom=10pt
]
\scriptsize

Please carefully observe this medical image and answer the following question: \\

Question: \{question\} 

Options:\\
{\setlength{\tabcolsep}{1.5pt}
\begin{tabular}{@{}ll}
A) & \{option\_Text\_A\} \\
B) & \{option\_Text\_B\} \\
C) & \{option\_Text\_C\} \\
D) & \{option\_Text\_D\} \\
E) & \{option\_Text\_E\} \\
\end{tabular}
}\\

Answer only with the option letter (A–E).

\end{tcolorbox}

\caption{Prompt of MCQ response generation.}
\end{table}

\subsection{Open-ended Response Generation Prompt}

\begin{table}[h]
\centering
\setlength{\abovecaptionskip}{3pt}

\begin{tcolorbox}[
    colback=gray!15,
    colframe=black,
    arc=1.3mm,
    boxrule=0.8pt,
    width=0.95\linewidth,
    left=10pt,
    right=10pt,
    top=10pt,
    bottom=10pt
]
\scriptsize

Please carefully observe this medical image and answer the following question: \\

Question: \{question\} \\

Think step by step, integrating both visual features and medical knowledge to reach your conclusion. Then provide the final answer to the question in one short sentence or a single medical term.\\

Output format (Must follow strictly):\\
Reasoning: <visual and diagnostic reasoning process>\\
Answer: <final answer to the question>

\end{tcolorbox}

\caption{Prompt of open-ended response generation.}
\end{table}

\newpage
\subsection{Open-ended Response Scoring Prompt}
\vspace{-1em}
\begin{table}[h]
\centering

\begin{tcolorbox}[
    colback=gray!15,
    colframe=black,
    arc=1.3mm,
    boxrule=0.8pt,
    width=0.95\linewidth,
    left=10pt,
    right=10pt,
    top=10pt,
    bottom=10pt
]
\scriptsize
You are a comprehensive and unbiased medical imaging evaluation assistant.\\[3pt]
You are given the following inputs:

0. [QUESTION]: \{question\}\\
- This is the **original diagnostic or descriptive question** posed to the model. It defines what the model is expected to describe, explain, or conclude about the image. It provides the evaluation context and should be considered when judging relevance and reasoning alignment.\\[3pt]
1. [IMAGE CAPTION]: \{image\_caption\}\\
- This is a **factual visual description** of the image content. It objectively states what is seen in the image, including structures, densities, shapes, boundaries, or textures. It serves as the visual ground truth for all vision-related evaluation.\\[3pt]
2. [GROUND TRUTH]: \{ground\_truth\}\\
- This is the **expert-verified reference interpretation** describing the clinically correct diagnosis or observation derived from the image. It represents the gold standard for semantic and factual correctness, not the visual features themselves.\\[3pt]
3. [MODEL REASONING]: \{model\_reasoning\}\\
This is the **model’s reasoning process** that explains how it interprets the visual information and arrives at its final answer. It may include descriptive, inferential, or diagnostic steps.\\[3pt]
4. [MODEL ANSWER]: \{model\_answer\}\\
- This is the **model’s final conclusion or diagnostic output**, which summarizes its interpretation of the image based on the reasoning process.\\[5pt]
You must evaluate the model output starting from Clarity and end with Ground Truth Consistency. Each aspect must be scored independently; do not let one aspect’s judgment influence the others.\\[2pt]
- The first two aspects (Language Clarity and Reasoning Coherence) are evaluated only on linguistic and logical quality, **not factual correctness**.\\
- The Vision Feature Accuracy aspect compares the model’s visual understanding against the **IMAGE CAPTION only**, **without considering medical correctness or ground truth content**.\\
- Only the last aspect (Ground Truth Consistency) compares reasoning and answer with the **GROUND TRUTH** for factual and semantic accuracy.

There are four independent aspects to evaluate:\\
1. Language Clarity (10\%)\\
2. Reasoning Coherence (10\%)\\
3. Vision Feature Accuracy (40\%)\\
4. Ground Truth Consistency (40\%)\\[5pt]
\textbf{Step 1 — Language Clarity (10\%)}\\
Evaluate whether [MODEL REASONING] and [MODEL ANSWER] are clear and unambiguous in expression, without internal contradictions or vague wording.\\
Scores:\\
0: Ambiguous, contradictory, or difficult to understand; key meaning is unclear.\\
0.5: Generally understandable but includes ambiguous terms, unclear phrasing, or minor inconsistencies.\\
1: Fully clear, precise, and internally consistent with no ambiguity.\\[4pt]
\textbf{Step 2 — Reasoning Coherence (10\%)}\\
Evaluate whether [MODEL REASONING] is logically coherent and consistent, ensuring that the reasoning flow from description to conclusion is smooth and medically reasonable.\\
Scores:\\
0: Clear logical errors or incoherent reasoning flow.\\
0.5: Mostly coherent but includes small logical gaps or weak causal links.\\
1: Fully coherent, consistent, and logically well-structured reasoning.\\[4pt]
\textbf{Step 3 — Vision Feature Accuracy (40\%)}\\
Evaluate whether [MODEL REASONING] and [MODEL ANSWER] correctly reflect the key visual features described in [IMAGE CAPTION], including structure, density, morphology, boundary, and texture.\\
Do not consider medical correctness or the ground truth content — only check the consistency with the image caption.\\
Scores:\\
0: No relevant visual features match the caption.\\
1: Only a few minor features match; most are inaccurate or missing.\\
2: Some correct features, but major structures or textures are misunderstood.\\
3: Mostly correct; captures main visual characteristics but includes noticeable or clinically relevant inaccuracies.\\
4: Completely correct; all key visual features match the caption with no meaningful omissions.\\[4pt]
\textbf{Step 4 — Ground Truth Consistency (40\%)}\\
Evaluate how well the combination of [MODEL REASONING] and [MODEL ANSWER] matches the [GROUND TRUTH] in terms of semantic meaning, factual correctness, and overall medical interpretation.\\
Scores:\\
0: Completely incorrect; reasoning and answer have no alignment with the ground truth.\\
1: Mostly incorrect; minor semantic overlap but conceptually different.\\
2: Partially correct; some overlap but with major factual or interpretive errors.\\
3: Mostly consistent; captures main meaning but with clear omissions or inaccuracies.\\
4: Perfectly consistent; reasoning and answer fully align with the ground truth meaning.\\[4pt]
Please output the results strictly in the following format (each score on a new line):\\
Language Clarity: [score]\\
Reasoning Coherence: [score]\\
Vision Feature Accuracy: [score]\\
Ground Truth Consistency: [score]
\end{tcolorbox}
\caption{Prompt of scoring the open-ended response.}
\end{table}

\clearpage

{
    \small
    \bibliographystyle{ieeenat_fullname}
    \bibliography{main}
}


\end{document}